\documentclass[preprint,12pt]{elsarticle}

\usepackage{amssymb}

\usepackage{graphicx}
\usepackage{soul}
\usepackage{multirow}
\usepackage{chemformula}
\usepackage{multirow}
\usepackage{colortbl}
\usepackage{adjustbox}
\usepackage{algorithm}
\usepackage{algpseudocode}
\usepackage{booktabs}
\usepackage{todonotes}
\usepackage{multirow}
\usepackage{tikz}
\usepackage{makecell}

\usepackage{graphicx}
\usepackage{xcolor}
\usepackage{caption}
\usepackage{subcaption}
\usepackage{lscape}
\usepackage{algorithm}
\usepackage{algpseudocode}
\usepackage{tikz}
\usetikzlibrary{positioning}

\RequirePackage[colorlinks,citecolor=black,urlcolor=black, bookmarks=false]{hyperref}

\usepackage{lineno}
\usepackage{lscape}

\begin{document}

\begin{frontmatter}



\title{Early Detection of Red Palm Weevil Infestations using Deep Learning Classification of Acoustic Signals}

\author[label1,label2]{Wadii Boulila\corref{cor1}}
\ead{wboulila@psu.edu.sa}
\cortext[cor1]{Corresponding author}

\author[label1]{Ayyub Alzahem}
\ead{aalzahem@psu.edu.sa}

\author[label1]{Anis Koubaa}
\ead{akoubaa@psu.edu.sa}

\author[label1,label3]{Bilel Benjdira}
\ead{bbenjdira@psu.edu.sa}

\author[label1]{Adel Ammar}
\ead{aammar@psu.edu.sa}

\address[label1]{Robotics and Internet-of-Things Laboratory, Prince Sultan University, Riyadh 12435, Saudi Arabia}
\address[label2]{RIADI Laboratory, National School of Computer Sciences, University of Manouba, Manouba, Manouba 2010, Tunisia}
\address[label3]{SE \& ICT Lab, LR18ES44, ENICarthage, University of Carthage, Tunis 1054, Tunisia}

\begin{abstract}
The Red Palm Weevil (RPW), also known as the palm weevil, is considered among the world's most damaging insect pests of palms. Current detection techniques include the detection of symptoms of RPW using visual or sound inspection and chemical detection of volatile signatures generated by infested palm trees. However, efficient detection of RPW diseases at an early stage is considered one of the most challenging issues for cultivating date palms. In this paper, an efficient approach to the early detection of RPW is proposed. The proposed approach is based on RPW sound activities being recorded and analyzed. The first step involves the conversion of sound data into images based on a selected set of features. The second step involves the combination of images from the same sound file but computed by different features into a single image. The third step involves the application of different Deep Learning (DL) techniques to classify resulting images into two classes: infested and not infested. Experimental results show good performances of the proposed approach for RPW detection using different DL techniques, namely MobileNetV2, ResNet50V2, ResNet152V2, VGG16, VGG19, DenseNet121, DenseNet201, Xception, and InceptionV3. The proposed approach outperformed existing techniques for public datasets.
\end{abstract}


\end{frontmatter}

\section{Introduction}


The red palm weevil (RPW), \emph{Rhynchophorus ferrugineus}, is a significant pest of palm trees, causing considerable damage to the tree and resulting in substantial financial losses for farmers. RPW is considered among the most destructive pests that threaten palms worldwide. Saudi Arabia provides about \$2 million contribution to support the global efforts to eradicate RPW \cite{rpwfao}. However, several limitations still need to be addressed, especially on the farm side. In most cases, farm owners did not have any automated process providing clear indicators about infested palm trees, which causes from 25 to 40\% loss of harvest due to unregulated processes \cite{el2010degradation}.

The management of the RPW has presented a significant challenge for entomologists since the beginning of modern agriculture, especially in the early detection of infested palms. Visual inspection of palms is the most adopted technique to locate infested palms. To detect palm diseases, farm owners would need to pay a significant amount of money to hire a palm tree infection expert for a whole year. Depending on the terrain and density, an expert can inspect around 200 palm trees daily. In many cases, there is up to 40\% loss of crop yield due to the inexperience of the farmworkers or unregulated processes. In addition, the problem of RPW detection becomes more difficult when it comes to early detection \cite{khan2022deep,hu2022uav}. Indeed, RPW is a hidden pest for most of its life stages, except for the adult stage. Hence, it is imperative to ensure proper measures for early RPW detection. RPW causes various morphological changes to the palm tree, such as the formation of holes in the trunk and leaves, wilting, and discoloration. Infested palm trees often have a foul smell, and their fronds turn yellow and droop. In the later stages of infestation, the crown of the palm tree collapses, leading to the death of the entire tree. However, these morphological changes are challenging to detect in the early stages of infestation. Therefore, using an alternative technique, such as sound data, becomes crucial in detecting the presence of the RPW. Using sound data provides a non-invasive and efficient method for early detection of RPW infestation in palm trees, making it a valuable tool for farm owners and professionals in the agricultural industry.

For these reasons, there is a pressing need to use advanced information technologies in date farming. Deep learning (DL) has recently shown great promise in various applications, from natural language processing and computer vision to healthcare and finance \cite{rehman2022novel,ben2022randomly,boulila2022hybrid}. DL models can learn complex patterns and relationships in data and make predictions or decisions based on that learned knowledge \cite{al2021novel,ben2022fusion}. These models can learn from large amounts of data and can improve their performance \cite{ghandorh2022semantic}. Due to these capabilities, DL is being used to solve many problems, including detecting the red palm weevil from sound data.

In this study, a novel approach for early disease detection is proposed. The proposed approach is a DL-based method for detecting infestation in palm trees using sound data. The process starts by recording sound data from palm trees and converting it into images by computing a set of selected features. The images are combined and classified using DL techniques to identify an infestation.

The main contributions of the proposed approach are summarized as follows:
\begin{itemize}
    \item We proposed to combine three types of audio feature extraction techniques: Constant Q Cepstral Coefficients (CQCC) \cite{su2023robust}, Mel-frequency Cepstral Coefficients (MFCC) \cite{ABBASKHAH2023105261}, and Bark Frequency Cepstral Coefficients (BFCC) \cite{gambhir2023end}. These features are combined to improve the classification accuracy of RPW infestation.
    \item We proposed an efficient method to detect early signs of RPW infestation, providing early intervention from farm owners and professionals.
    \item Several experiments are conducted using three public datasets: TreeVibe \cite{smartcities4010017}, Environmental Sound Classification (ESC) \cite{inik2023cnn}. Results show that the proposed approach provides excellent performances compared to existing related works.
\end{itemize}

The rest of the paper is organized as follows. Section 2 reviews relevant literature on the detection of palm tree infestation, highlighting key findings and gaps in the existing research. Section 3 details the study area, dataset description, and data collection. Section 4 describes the research design and analysis procedures used in the study. Results are presented in Section 5. In this section, the findings of the proposed study, including a comparison with existing techniques for public datasets. Section 6 interprets the results of the study in the context of the existing literature and provides insights into the implications of the findings. Finally, the main findings of the proposed study, their implications, and recommendations for future research are summarized in Section 7.

\section{Related works}

Recently, the application of new and modern agricultural techniques has proved its efficiency in many agriculture-related fields \cite{roopaei2017cloud,bu2019smart,yang2021survey,sinha2022recent}. In the context of our study, we are interested in reviewing the use of modern technologies in the case of early disease detection, which is among the top priorities of farm stakeholders since it will ensure healthy farms and high-quality crops. In the literature, several works aim to use modern technologies to ensure the early detection of plant infestation \cite{hu2022detection,haridasan2023deep,mallick2023deep,yu2023inception,zhu2023knowledge}. However, this research area still needs more investigation, especially for palm trees \cite{ferreira2021accurate,goldshtein2022analyzing,putra2022oil}. In this paper, we have reviewed research works closely aligned with our research endeavors, which is the early detection of RPW infestation using acoustic signals. Koubaa et al. in \cite{koubaa2020smart} proposed an IoT-based framework for red palm weevil early detection. The proposed framework allows the detection of red palm weevils by using smart agriculture sensors and collecting data using accelerometer sensors. Signal processing and probability methods are developed to identify the activity of the red palm weevil. Pinhas et al. \cite{PINHAS2008131} developed a mathematical method to automatically detect the acoustic signals of RPW larvae in palm offshoots using off-the-shelf recording devices. They utilized techniques similar to those used in speech recognition, such as vector quantization and Gaussian mixture modeling, to achieve detection ratios of 98.9\%. This method could be used to monitor the trade and transportation of offshoots, helping to prevent the spread of this pest. In this paper \cite{parvathy2021convolutional}, a Convolutional Autoencoder based Deep Learning model is proposed for identifying Red Palm Weevil (RPW) signals from background noise. The model uses Mel spectrogram images of RPW acoustic activities as the training data, achieving a classification accuracy of 95.85\% on the test dataset composed of normal RPW acoustic emissions as well as anomalous acoustic samples. The proposed method is found to be highly efficient for the identification of RPW signals.
Mohamed et al. \cite{9686081} proposed a method based on recording RPW sounds during different activity states, extracting important features in the sounds using the MFCC technique, and using artificial neural networks to automatically detect if RPW larvae exists. The recorded sounds were transformed into a spectrogram and analyzed with a CNN model, achieving 99.2\% accuracy using AlexNet \cite{alom2018history}. This AI-based approach could help to detect RPW larvae early, potentially preventing the spread of this pest.
In \cite{s21051592}, the authors introduced the combination of machine learning and fiber optic distributed acoustic sensing (DAS) techniques as a solution for the early detection of RPW in vast farms. They trained a fully-connected artificial neural network and a convolutional neural network using experimental time- and frequency-domain data provided by the fiber optic DAS system. The models can efficiently recognize healthy and infested trees with high classification accuracy values (99.9\% by ANN with temporal data and 99.7\% by CNN with spectral data, in reasonable noise conditions). This work paves the way for deploying the high-efficiency and cost-effective fiber optic DAS to monitor RPW in open-air and large-scale farms containing thousands of trees.
In \cite{s22176491}, the authors integrated optical fiber DAS and machine learning (ML) for the early detection of true weevil larvae less than three weeks old. They recorded temporal and spectral data using the DAS system and processed it with a 100–800 Hz filter to train CNN models that distinguish between "infested" and "healthy" signals with a classification accuracy of ~97\%. They also introduced a strict ML-based classification approach to improving the false alarm performance metric of the system. This approach could help to detect RPW early, preventing the spread of this harmful pest.
In \cite{ESMAILKARAR20225309}, the authors proposed an IoT-based system for early detection of RPW larvae in date palms using a modified mixed depthwise convolutional network (MixConvNet) as a deep learning classifier. The proposed system was tested on the public TreeVibes dataset and achieved an accuracy score of 97.38\%. The proposed system is considered a future milestone for practical implementation in the field.

Table \ref{tb:related_works} summarizes the discussed related works. As shown in the related works, most authors rely on spectral data in their research. In this paper, we present an innovative way to detect RPW based on the combination of features extracted from the collected sound data. 

\begin{landscape}
\begin{table}
    \scriptsize
    \centering
    \caption{Comparison of recent research works related to RPW detection using audio signals .}
    \begin{tabular}{p{1cm}p{1cm}p{2cm}p{3cm}p{7cm}p{3cm}p{2cm}}
        \multicolumn{1}{c}{\textbf{Ref.}} & 
        \multicolumn{1}{c}{\textbf{Year}} &
        \multicolumn{1}{c}{\textbf{Algorithm}} & 
        \multicolumn{1}{c}{\textbf{Data Type}} & 
        \multicolumn{1}{c}{\textbf{Steps}} & 
        \multicolumn{1}{c}{\textbf{Dataset}} &  
        \multicolumn{1}{c}{\textbf{Accuracy}} \\ \hline
        \multirow{6.5}{1cm}{\centering \cite{PINHAS2008131}} & 
        \multirow{6.5}{1cm}{\centering 2008} & 
        \multirow{6.5}{2cm}{\centering GMM \\ VQ} & 
        \multirow{6.5}{3cm}{\centering Spectral} 
        & 
        \vspace{0.2cm}
        $\bullet$ Record the RPW using off the helf device\newline
        $\bullet$ Manual data labeling\newline
        $\bullet$ Calculate noise threshold\newline
        $\bullet$ Counting the mismatched labels
        \vspace{0.4cm}
        & 
        \multirow{6.5}{3cm}{\centering -} & 
        \multirow{6.5}{2cm}{\centering 0.988 \\ 0.989} \\ \hline
        \multirow{6.5}{1cm}{\centering \cite{parvathy2021convolutional}} & 
        \multirow{6.5}{1cm}{\centering 2021} & 
        \multirow{6.5}{2cm}{\centering CAE} & 
        \multirow{6.5}{3cm}{\centering Mel Spectrograms} 
        & 
        \vspace{0.2cm}
        $\bullet$ Capture audio signals \newline
        $\bullet$ Extract features using Mel Spectrograms algorithm \newline
        $\bullet$ Convert the features to images \newline
        $\bullet$ Train CAE on the images 
        \vspace{0.4cm}
        & 
        \multirow{6.5}{3cm}{\centering -} & 
        \multirow{6.5}{2cm}{\centering  0.958} \\ \hline
        \multirow{6}{1cm}{\centering \cite{9686081}} & 
        \multirow{6}{1cm}{\centering  2021} & 
        \multirow{6}{2cm}{\centering  AlexNet} & 
        \multirow{6}{3cm}{\centering  MFCC} 
        & 
        \vspace{0.2cm}
        $\bullet$ Extract features using MFCC algorithm \newline
        $\bullet$ Convert the features to images \newline
        $\bullet$ Train AlexNet on the images 
        \vspace{0.4cm}
        & 
        \multirow{6}{3cm}{\centering  -} & 
        \multirow{6}{2cm}{\centering  0.992} \\ \hline
        \multirow{6.5}{1cm}{\centering \cite{s21051592}} & 
        \multirow{6.5}{1cm}{\centering  2021} & 
        \multirow{6.5}{2cm}{\centering  ANN \\ CNN} & 
        \multirow{6.5}{3cm}{\centering  Temporal \\ Spectral} 
        & 
        \vspace{0.2cm}
        $\bullet$ Install fiber optic DAS system on palm trees \newline
        $\bullet$ Capture audio signals \newline
        $\bullet$ Train ANN and CNN with DAS data \newline
        $\bullet$ Deploy DAS system in large farms 
        \vspace{0.4cm}
        & 
        \multirow{6.5}{3cm}{\centering  -} & 
        \multirow{6.5}{2cm}{\centering  0.999 \\ 0.997} \\ \hline
        \multirow{6.5}{1cm}{\centering \cite{s22176491}} & 
        \multirow{6.5}{1cm}{\centering  2022} & 
        \multirow{6.5}{2cm}{\centering  CNN} & 
        \multirow{6.5}{3cm}{\centering  Temporal \\ Spectral} 
        & 
        \vspace{0.2cm}
        $\bullet$ Record the sound of RPW using DAS \newline
        $\bullet$ Transform it into temporal and spectral data \newline
        $\bullet$ Train CNN on the data separately \newline
        $\bullet$ Merge CNN models 
        \vspace{0.4cm}
        & 
        \multirow{6.5}{3cm}{\centering  -} & 
        \multirow{6.5}{2cm}{\centering  0.970 \\ 0.971} \\ \hline
        \multirow{6.5}{1cm}{\centering \cite{ESMAILKARAR20225309}} & 
        \multirow{6.5}{1cm}{\centering 2022} & 
        \multirow{6.5}{2cm}{\centering CNN} & 
        \multirow{6.5}{3cm}{\centering Spectral} 
        & 
        \vspace{0.2cm}
        $\bullet$ Extract spectral features \newline
        $\bullet$ Convert the features to images \newline 
        $\bullet$ Train MixConvNet on the images \newline
        $\bullet$ Compare performance with other models 
        \vspace{0.4cm}
        & 
        \multirow{6.5}{3cm}{\centering TreeVibes} & 
        \multirow{6.5}{2cm}{\centering 0.973} \\ \hline
    \end{tabular}
    \label{tb:related_works}
\end{table}
\end{landscape}


\section{Dataset collection and description}

Saudi Arabia is the second-largest palm date producer. Al-Ahssa region in Saudi Arabia was taken as a case study to test the performance of the proposed approach. The dataset used in this study originally is a set of recordings. The recordings were taken from date palm trees using remote vibroacoustic surveillance of trees. The device\footnote{\url{http://www.insectronics.net/}.} used in this study is placed on the palm tree's trunk and records the vibrations created by the weevils as they chew and move inside the tree.


The reading points were marked on each date palm trunk (one meter above the ground level). Then drill holes (30 degrees) were made using a drill machine equipped with a brad point drill-bit (diameter, 8 mm) for a 35 cm depth inside the palm tree. A single hole was made on the site of the trunk, and an acoustic device covered the whole tree for RPW sound detection. For the reading, the acoustic device has a probe (waveguide) (35 cm in length) that is inserted inside the tree embedded with a recording machine. The waveguide was inserted into the hole of the date palm trunk that was previously drilled. The machine recording is connected to headphones that enable the user to assess the sound signal produced by RPW activities. This device provides the data which enables the movement from the device and the server/cloud system. The RPW device can detect the sound signal inside the date palm tree within a sphere of 1.5-2 meters radius \cite{smartcities4010017}. In this experiment, the parameter was set up to five minutes of repeat period and 20 seconds of recording times. Twenty seconds of sound data are contained in each tree recording daily. This record protocol was done in all date palm trees. Time stamps of each recording were also provided.

In order to better understand the dataset, Table \ref{tab:dataset} shows the number of recordings per class.

\begin{table}[h]
\centering
 \caption{Data description.}
 \label{tab:dataset}
   \begin{tabular}{@{}|c|c|@{}}
\hline
\textbf{Class} & \textbf{Number of samples}  \\ \hline
Infested & 531  \\ \hline
Not Infested & 575  \\ \hline
\end{tabular}
\end{table}

After collecting the data, several features are computed based on the sound dataset. Then, these features are combined to better describe the content of the data. The process of feature extraction and collaboration is described in the following section.

\section{Proposed Approach}

\begin{figure}
\centering
\includegraphics[width=\textwidth]{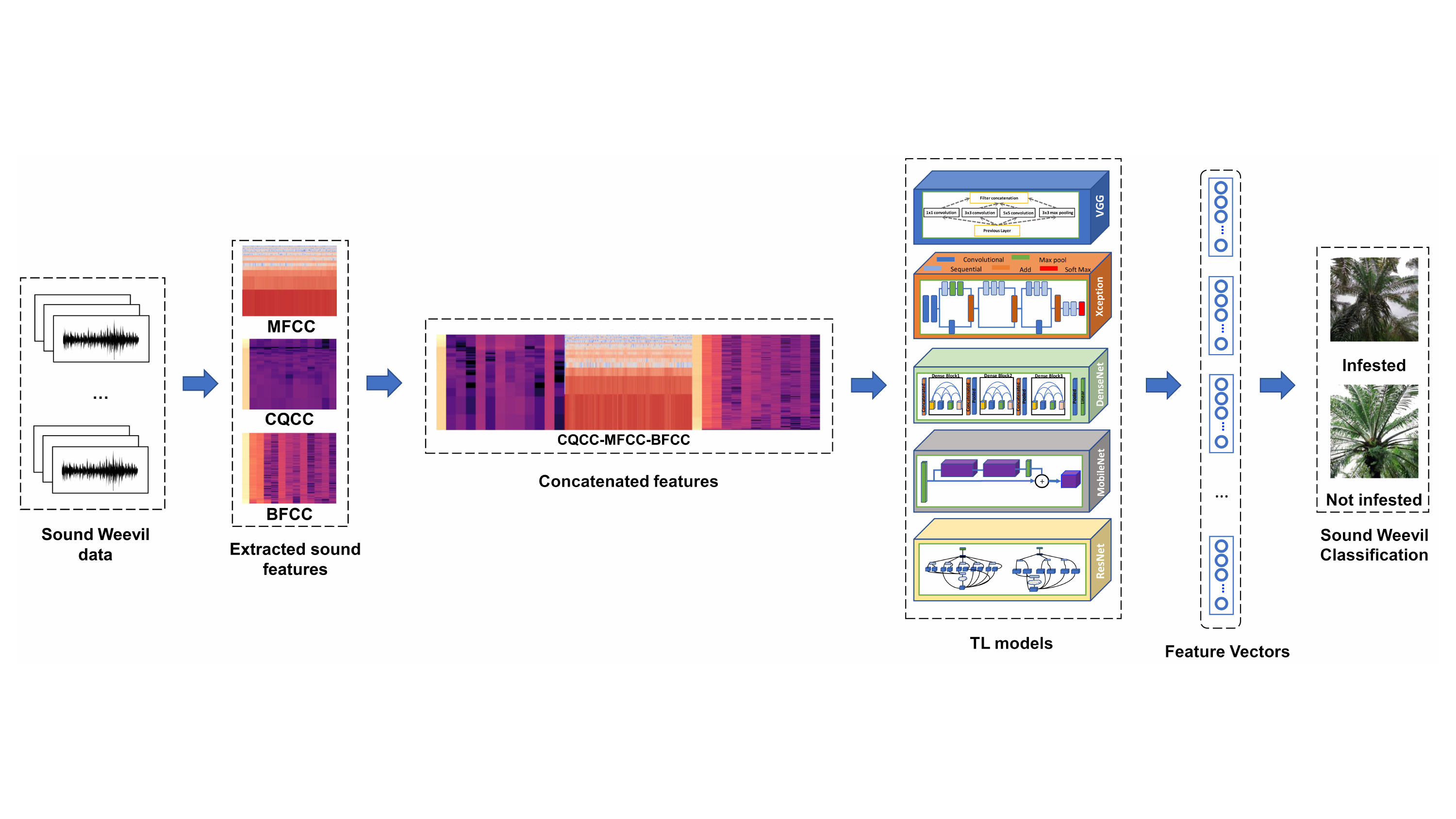}
\caption{Architecture of the proposed approach.} 
\label{fig:approach}
\end{figure}

The goal of this study is to detect RPW infestation from sound data acquired from palm trees. Figure \ref{fig:approach} describes the architecture of the proposed approach. The process starts by converting sound data into images by computing a set of selected features. The second step will combine images of the same sound file (computed by different features) into a single image. The third step involves applying different DL techniques to classify resulting images into two classes: infested and not infested.

\subsection{Feature extraction from sound Weevil dataset}

We used CQCC, MFCC, and BFCC feature extractions to convert one sound file into three images, one for each feature. 
 
To train DL models, extracting useful features from recorded sound data is required. In literature, several features have been investigated \cite{chen2021deep}. A waveform can be converted to a spectrogram, which is a visual representation of all frequencies over time. Many features can be computed on sound data. In this study, we will consider MFCC, CQCC, and BFCC.

\subsubsection{Mel-Frequency Cepstral Coefficients (MFCC)}

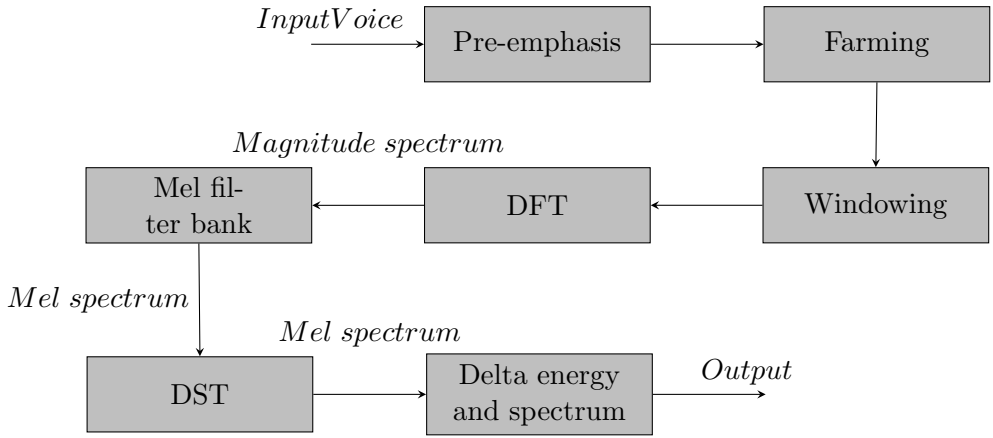
\begin{figure} [h]

\begin{tikzpicture}
\small
\node [fill=none] (block1) at (0,0) {};
\node [draw,
    fill=lightgray,
    minimum width=3cm,
    minimum height=1cm,
    right=1.5cm of block1
]  (block2)[text width=2.5cm,align=center] {Pre-emphasis};

\node [draw,
    fill=lightgray, 
    minimum width=3cm, 
    minimum height=1cm,
    right=1.5cm of block2
]  (block3)[text width=2.5cm,align=center] {Farming};

\node [draw,
    fill=lightgray, 
    minimum width=3cm, 
    minimum height=1cm, 
    below right= 1.5cm and 6cm of block1
]  (block4)[text width=2.5cm,align=center] {Windowing};

\node [draw,
    fill=lightgray, 
    minimum width=3cm, 
    minimum height=1cm, 
    below right= 1.5cm and 1.5cm of block1
]  (block5)[text width=2.5cm,align=center] {DFT};

\node [draw,
    fill=lightgray, 
    minimum width=3cm, 
    minimum height=1cm, 
    below right= 1.5cm and -3cm of block1
]  (block6)[text width=2.5cm,align=center] {Mel filter bank};

\node [draw,
    fill=lightgray, 
    minimum width=3cm, 
    minimum height=1cm, 
    below right= 1.5cm and -3cm of block6
]  (block7)[text width=2.5cm,align=center] {DST};

\node [draw,
    fill=lightgray, 
    minimum width=3cm, 
    minimum height=1cm, 
    right=1.5cm of block7
]  (block8)[text width=2.5cm,align=center] {Delta energy and spectrum};

\node [
    fill=none, 
    right=1.5cm of block8
]  (block9)[text width=2.5cm,align=center] {};

\draw[-stealth] (block1.east) --  (block2.west)
    node[midway,above, xshift=-0.5cm]{$Input Voice$};
 
\draw[-stealth] (block2.east) -- (block3.west) 
    node[midway,above, yshift=0.5cm]{};
 
\draw[-stealth] (block3.south) -- (block4.north)
    node[midway,right]{};

\draw[-stealth] (block4.west) -- (block5.east) 
    node[midway,above, yshift=0.5cm]{};

\draw[-stealth] (block5.west) -- (block6.east) 
    node[midway,above, yshift=0.5cm] {$Magnitude\;spectrum$} ;

\draw[-stealth] (block6.south) -- (block7.north)
    node[midway,left]{$Mel\;spectrum$};
    
\draw[-stealth] (block7.east) -- (block8.west)
    node[midway,above, yshift=0.5cm]{$Mel\;spectrum$};
 
\draw[-stealth] (block8.east) -- (block9.west) 
    node[midway,above, xshift=0.5cm]{$Output$};
\end{tikzpicture}
\caption{MFCC Block Diagram. \label{fig:mfcc}}
\end{figure}

The cepstrum of a signal contains information about the rate of change in its spectral bands. A cepstrum is just a spectrum of the log of the temporal signal's spectrum. MFCCs are coefficients that make up the mel-frequency cepstrum \cite{zheng2001comparison}. The cepstrum communicates values that constitute a sound's formants (a distinguishing component of its quality) and timbre. As a result, MFCCs are very useful in DL models. Figure \ref{fig:mfcc} depicts the block diagram of MFCC \cite{ibrahim2008quranic}.

\subsubsection{Constant Q Cepstral Coefficients (CQCC)}

Combining the constant Q transform (CQT) and conventional cepstral processing results in the CQCC. CQT's design is appropriate for musical representation and transforms a data series to the frequency domain \cite{brown1991calculation}. It is a time-frequency analysis that perceptually motivated more than The short-term Fourier transform (STFT) \cite{allen1977short}. The ratio of the center frequency $f_k$ to the bandwidth $\delta f$ is known as the $Q$ factor, which measures the selectivity of each filter.
\begin{equation}
  Q = \frac{f_k}{\delta f}
\end{equation}
The window function of the STFT is coupled to the constant bandwidth of each filter. Since all filters have the same absolute bandwidth $f$, the $Q$ factor increases as one moves from low to high frequencies. Contrarily, the Q factor of the human sensory system is known to remain about constant between 500Hz and 20kHz (Moore, 2003). So, from a perceptual perspective, the STFT might not always be the optimum choice for the time-frequency analysis of speech signals.\\
The transform can be represented as a succession of filters $f_k$, each one $k-th$ having a spectral width $\delta f_k$ that is a multiple of the width of the filter before it: 
\begin{equation}
  \delta f_k = 2^{1/n} \cdot \delta f_{k-1} = (2^{1/n})^k = \delta f_{min}
  \label{eq:fk}
\end{equation}
Where $n$ is the number of filters per octave, $f_{min}$ is the center frequency of the lowest filter, and $delta f_k$ is the bandwidth of the $k-th$ filter.\\
For CQT computation, the discrete time domain signal $x(n)$ has the following CQT $X^{CQ}(k, n)$ definition: 
\begin{equation}
   X^{CQ}(k, n) = \sum_{j=n-\lfloor N_k/2 \rfloor}^{n+\lfloor N_k/2 \rfloor} x(j) a^*_k (j-n+\frac{N_k}{2})
  \label{eq:cqt}
\end{equation}
Where a $a^*_k(n)$ is the complex number of the basis functions $a_k(n)$, K is the frequency bin index, $k$ = 1, 2,...,$K$ and $N_k$ are variable window lengths. The notation $\lfloor \cdot \rfloor$ denotes the floor function. The $a_k(n)$ functions are diverse time-frequency elements, defined as: 
\begin{equation}
  a_k(n) = \frac{1}{C} (\frac{n}{N_k}) \: exp[i(2\pi n \frac{f_k}{f_s} + \phi_k)]
  \label{eq:ak}
\end{equation}
Where $k$ is a phase offset, $f_k$ is the bin $k$'s center frequency, and $f_s$ is the sampling rate. The scaling factor $C$ can be found by using:
\begin{equation}
  C = \sum_{l=-\lfloor N_k/2 \rfloor}^{\lfloor N_k/2 \rfloor} w\Biggl(\frac{l+N_k/2}{N_k}\Biggl)
  \label{eq:c}
\end{equation}
Where $w(x)$ represents a window function. The equation \ref{eq:cqcc} then allows for the more-or-less traditional extraction of CQCC. 
\begin{equation}
  CQCC(p) = \sum_{l=1}^{L} log|X^{CQ}(l)|^2 \: cos\bigg[\frac{p(l-\frac{1}{2})\pi}{L}\bigg]
  \label{eq:cqcc}
\end{equation}
Where $l$ are the recently resampled frequency bins and $p$ = 0, 1,..., $L$ - 1.

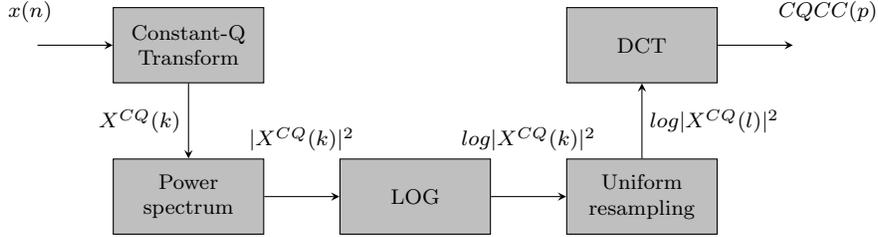
\begin{figure} [h]

\label{fig:cqcc}
\begin{tikzpicture}
\centering
\scriptsize

\node [
    fill=none,
    minimum width=1.5cm,
    minimum height=0.0cm,
]  (block1) at (0,0) {};

\node [draw,
    fill=lightgray,
    minimum width=2cm,
    minimum height=1cm,
    right=1cm of block1
]  (block2)[text width=1.6cm,align=center] {Constant-Q\\Transform};

\node [draw,
    fill=lightgray, 
    minimum width=2cm, 
    minimum height=1cm,
    below=1cm of block2
] (block3)[text width=1.26cm,align=center] {Power spectrum};

\node [draw,
    fill=lightgray, 
    minimum width=2cm, 
    minimum height=1cm, 
    right=1cm of block3
]  (block4)[text width=1.26cm,align=center] {LOG};

\node [draw,
    fill=lightgray, 
    minimum width=2cm, 
    minimum height=1cm, 
    right=1cm of block4
]  (block5) [text width=1.5cm,align=center] {Uniform\\resampling};

\node [draw,
    fill=lightgray, 
    minimum width=2cm, 
    minimum height=1cm, 
    above=1cm of block5
]  (block6)[text width=1.26cm,align=center] {DCT};
\node [
    fill=none, 
    minimum width=2cm, 
    minimum height=1cm, 
    right=1cm of block6
]  (block7)[text width=1.26cm,align=center] {};

\draw[-stealth] (block1.east) -- (block2.west)
    node[midway, above left=0.2cm and 0.2cm]{$x(n)$};
 
\draw[-stealth] (block2.south) -- (block3.north) 
    node[midway,left]{$X^{CQ}(k)$};
 
\draw[-stealth] (block3.east) -- (block4.west)
    node[midway,above,above=0.5cm]{$|X^{CQ}(k)|^2$};

\draw[-stealth] (block4.east) -- (block5.west) 
    node[midway,above,above=0.5cm]{$log|X^{CQ}(k)|^2$};

\draw[-stealth] (block5.north) -- (block6.south) 
    node[midway,right] {$log|X^{CQ}(l)|^2$} ;

\draw[-stealth] (block6.east) -- (block7.west)
    node[midway,above right=0.2cm and 0.2cm]{$CQCC(p)$};

\end{tikzpicture}
\caption{CQCC Block Diagram.}
\end{figure}

\subsubsection{Bark Frequency Cepstral Coefficients (BFCC)}

BFCC is a feature extraction technique used in speech and audio processing. It is based on the Bark scale, which is a non-linear frequency scale that approximates the human auditory system's response to different frequency bands. The Bark scale is defined as:

\begin{equation}
    z = 26.81*(f/1000) - 0.53 * (f/1000)^2 + (4.5*10^-6) * (f/1000)^3
  \label{eq:bfcc}
\end{equation}

Where $f$ is the frequency in Hz, the BFCCs are calculated by first converting the audio signal into the Bark scale, applying the above formula on the frequency axis of the spectrogram of the audio signal, then taking the cepstral coefficients of the resulting Bark-scale spectrogram. The cepstral coefficients are obtained by applying a Discrete Cosine Transform (DCT) on the log-magnitude of the Bark-scale spectrogram. This process is known as Cepstral Analysis. The result is a set of coefficients that represents the audio signal in the Bark frequency scale. These coefficients are often used as input to machine learning models for tasks such as speech recognition and music classification.

\subsection{Feature combination}

DL techniques have the ability to learn representations of data using multiple abstraction levels. The feature combination aims to take advantage of multiple features to construct a robust discriminator of red weevil sound. Each computed feature determines a specific aspect of the audio signal's temporal or spectral content. Combining these features will help the DL techniques discriminate between infested and not infested palms. Using the early mentioned features, DL models can be trained separately using only one feature at a time or combined features.

Algorithm \ref{algo:combine} describes the detailed process of combining the extracted features using our selected feature extractions CQCC, MFCC, and BFCC, where lines 2 to 4 specify the static variables such as data paths and feature extractions. Lines 6 and 7 are for audio file loading based on the current file name on the loop at line 5. After loading our audio file, in lines 10 to 12, we extract the audio features and convert it to an image without any axis. In line 13, we append the generated image in a temporary list that clears itself each time after loading an audio file. After extracting the audio feature and appending the images to our temporary list, lines 15 to 17 combine the generated images and save them as one image with the same name as the audio file. This process will be repeated on all the audio files in the target folder.

\begin{algorithm}[H]
\caption{: Combine multiple feature extraction methods}
\label{algo:combine}
\begin{algorithmic}[1]
    \State{\textbf{Begin}}
        \State{$audioFolderPath \leftarrow /home/data/audio$}
        \State{$imagesFolderPath \leftarrow /home/data/images$}
        \State{$exteactionMethods \leftarrow [cqcc, mfcc, bfcc]$} \Comment{specify the extraction methods}
        \For{\textbf{each} $fileName \in audioFolderPath$ } \Comment{loop over all files in the audio folder}
                \State{$audioFilePath \leftarrow joinPath(audioFolderPath, fileName)$}
                \State{$audio = loadAudio(audioFilePath)$}
                \State{$tempImagesList \leftarrow \varnothing$}
            \For{\textbf{each} $method \in exteactionMethods$ } 
                \State{$features \leftarrow method.extract(audio)$} \Comment{use the current algorithm for the extraction}
                \State{$figure = createFigure()$} \Comment{create empty figure}
                \State{$figure.display(features)$} \Comment{display the extracted features on the figure}
                \State{$tempImagesList.append(figure)$} \Comment{save the features image temporarily}
            \EndFor
            \State{$combinedImage \leftarrow concat(tempImagesList)$} \Comment{concatenate features images}
            \State{$imageFilePath \leftarrow joinPath(imagesFolderPath, fileName)$}
            \State{$save(imageFilePath, combinedImage)$} 
        \EndFor
    \State{\textbf{End}}
\end{algorithmic}
\end{algorithm}

 \subsection{DL classification of sound Weevil}

We applied several DL techniques, namely MobileNetV2, ResNet50V2, ResNet152V2, VGG16, VGG19, DenseNet121, DenseNet201, Xception, and InceptionV3 on the generated Red Palm Weevil dataset. We used TensorFlow and Keras frameworks to train the listed TL models on 200 epochs, 0.0001 learning rate, and RMSprop optimizer. The dataset has been split into three sets, training set (0.8), validation set (0.1), and test set (0.1), to ensure preventing any results coming from overfitting.

These techniques are used to extract features from an audio dataset, concatenate the CQCC, MFCC, and BFCC images of each audio file together, and train models to classify the audio samples. CQCC, MFCC, and BFCC are feature extraction techniques commonly used in audio processing to extract different audio characteristics. By combining these features in the form of images and training CNNs on them, the models can learn and identify complex patterns in the audio.

These specific CNNs are chosen based on their performance and efficiency in the image and audio classification tasks. MobileNetV2 and Xception are lightweight models suitable for mobile and embedded vision applications, while ResNet50V2 and ResNet152V2 are variants of the ResNet architecture that improve accuracy by using residual connections and bottleneck layers. VGG16, VGG19, DenseNet121, DenseNet201, and InceptionV3 are also widely used CNNs that have been proven to be effective in various image and audio classification tasks. RMSprop optimizer was used for all training experiments as it is particularly well-suited for training deep neural networks with a large number of parameters.

Using a variety of CNNs in this study demonstrates the versatility and applicability of the proposed approach in various audio classification tasks and with different types of CNN architectures. The combination of CQCC, MFCC, and BFCC features with various CNNs allows the extraction of different audio characteristics and enhances the models' performance.

\section{Results}

\subsection{Feature extraction from the sound Weevil dataset}

\begin{figure}[H]
\centering
\includegraphics[width=\textwidth]{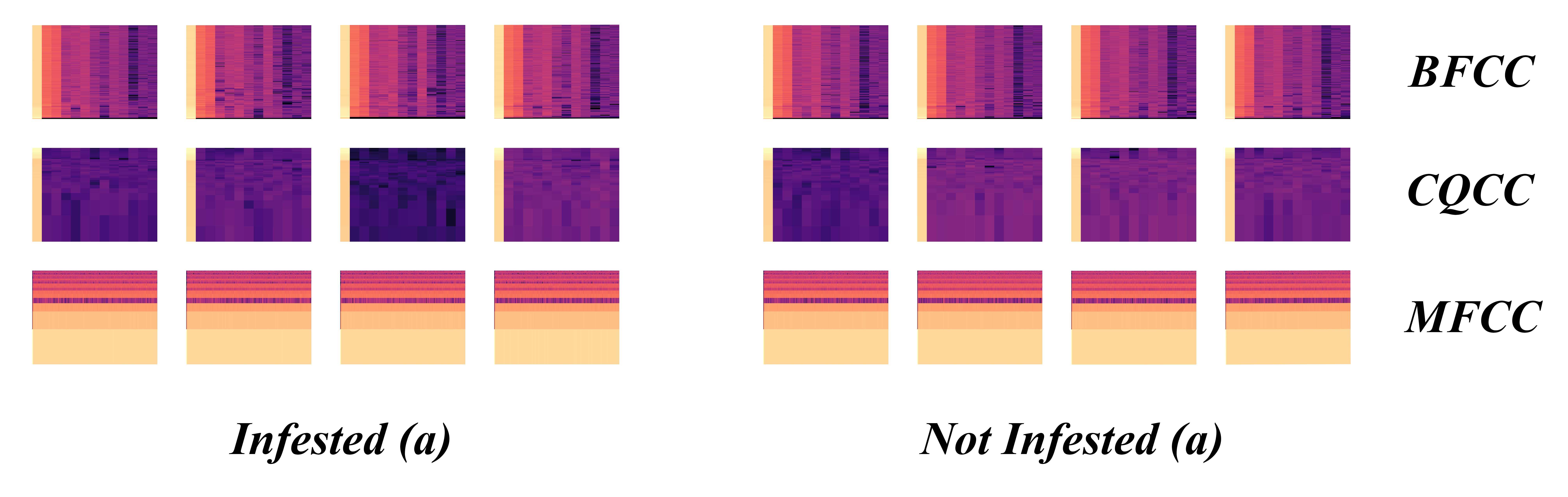}
\caption{The resulting images for each class.} 
\label{fig:samples}
\end{figure}

Feature extraction is a process in which characteristics or features of an audio signal are extracted and represented in a more manageable and informative way. It is crucial in many audio processing tasks, such as speech recognition, music classification, and sound event detection. The extracted features serve as input to machine learning models, which use them to make predictions or decisions.
There are many different feature extraction techniques that can be used, each with its own strengths and weaknesses. The choice of feature extraction method depends on the specific task and the characteristics of the audio signal.
Figure \ref{fig:samples} illustrates the use of three different feature extraction techniques, namely CQCC,  MFCC, and BFCC, on a dataset of audio files labeled as "infested" or "not infested." The figure contains four samples from each class, and the output of the three feature extraction techniques is displayed for each sample.

\subsection{Red Palm Weevil detection}

In this section, we present the validation accuracy results of transfer learning techniques with nine backbone models, namely MobileNetV2\cite{mobilenetv2}, ResNet50V2\cite{resnet50v2}, ResNet152V2\cite{resnet152v2}, VGG16\cite{vgg16}, VGG19\cite{vgg19}, DenseNet121\cite{densenet121}, DenseNet201\cite{densenet201}, Xception\cite{xception}, and InceptionV3\cite{inceptionv3},  for each feature extracted from the sound Weevil dataset as well as the combination of these features. The primary purpose of applying different DL techniques is to check the performance of the combination of selected sound features in detecting the RPW infestation.

Table \ref{table:backbones} provides a comparison of the performance of nine different machine learning models: MobileNetV2, ResNet50V2, ResNet152V2, VGG16, VGG19, DenseNet121, DenseNet201, Xception, and InceptionV3. These models are compared based on their top-1 accuracy (Accuracy@1), top-5 accuracy (Accuracy@5), the number of parameters (in millions), and their computational complexity in GFLOPS (Giga Floating Point Operations Per Second). The accuracies are evaluated on the ImageNet dataset and represent the model's ability to correctly identify the true class out of its top 1 and top 5 predictions, respectively. The number of parameters in the model represents the model's size and capacity, and GFLOPS is a measure of the computational complexity of the model, with higher values indicating more computation is required. The exact values for each model will need to be filled in according to the specific experiment results.

\begin{table}[h!]
\scriptsize
\centering
\begin{tabular}{|c|c|c|c|c|}
\hline
\textbf{Model} & \textbf{Accuracy@1} & \textbf{Accuracy@5} & \textbf{Parameters (in millions)} & \textbf{GFLOPS} \\
\hline 
MobileNetV2 \cite{mobilenetv2} & 72.154 & 90.822 & 3.5M & 0.3 \\
ResNet50V2 \cite{resnet50v2} & 80.858 & 95.434 & 25.6M & 4.09 \\
ResNet152V2 \cite{resnet152v2} & 82.284 & 96.002 & 60.2M & 11.51 \\
VGG16 \cite{vgg16} & 71.592 & 90.382 & 138.4M & 15.47 \\
VGG19 \cite{vgg19} & 72.376 & 90.876 & 143.7M & 19.63 \\
DenseNet121 \cite{densenet121} & 74.434 & 91.972 & 8.0M & 2.83 \\
DenseNet201 \cite{densenet201} & 76.896 & 93.37 & 20.0M & 4.29 \\
Xception \cite{xception} & 79.0 & 94.5 & 22.9M & - \\
InceptionV3 \cite{inceptionv3} & 77.294 & 93.45 & 27.2M & 5.71 \\
\hline
\end{tabular}
\caption{Comparative analysis of MobileNetV2, ResNet50V2, ResNet152V2, VGG16, VGG19, DenseNet121, DenseNet201, Xception, and InceptionV3 models on top-1 accuracy, top-5 accuracy, number of parameters, and GFLOPS using the ImageNet dataset.}
\label{table:backbones}
\end{table}

Table \ref{tb:single_feat_results} shows the accuracy and loss values for the classification experiments using the mentioned models. We used single-feature extraction algorithms to compare the classification performances of the considered DL models. The features depicted in the table are LFCC, CQCC, MFCC, Chroma, GFCC, BFCC, Mel Spectro, and Spectral Centroid. The highest accuracy was achieved by the models trained on the CQCC feature set, with a maximum accuracy of 1.000 using the DenseNet121 model. Achieving 100\% accuracy using only one feature, CQCC, is not sufficient because we aim to achieve the best accuracy regardless of the CNN architecture used. Therefore, we used multiple feature combinations to ensure high accuracy without relying on any specific CNN architecture. On the other hand, the lowest accuracy was achieved by the models trained on the MFCC feature, with a minimum accuracy of 52\% using the ResNet152V2 model.

In Table \ref{tb:multi_feat_results}, several experiments have been conducted with multiple feature combinations, including CQCC-MFCC-BFCC, MFCC-Chroma-Mel Spectro, MFCC-Chroma-LFCC, GFCC-BFCC-CQCC, GFCC-CQCC-LFCC, and Spectro-MFCC-BFCC. The accuracy values in the table ranged from 70\% to 100\%, indicating that the models performed very well on the test data. The loss values were all relatively low, ranging from 0.000 to 9.836. A lower loss value indicates that the model is better at predicting the target variable. It is interesting to note that all models trained on the CQCC-MFCC-BFCC feature set achieved perfect accuracy, while some of the models trained on other feature sets achieved lower accuracy values. The ResNet152V2 and DenseNet201 models performed very well across all feature sets. Table \ref{tb:multi_feat_results} highlights that the choice of a specific audio feature set has a significant impact on the accuracy of the models. It also suggests that the choice of model architecture is important for achieving good accuracy in the audio classification task.

\begin{table}[H]
    \scriptsize
    \centering
    \caption{Single feature experimental results.}
    \bgroup
    \def\arraystretch{1.5}%
    \begin{tabular}{|l|l|l|l|l|l|l|l|} 
        \hline
        \textbf{Features} & \textbf{Model} & \textbf{Accuracy} & \textbf{Loss} & \textbf{Features} & \textbf{Model} & \textbf{Accuracy} & \textbf{Loss}  \\ \hline
        \multirow{9}{*}{\centering LFCC} 
        & MobileNetV2 & 0.843 & 9.715 
        & \multirow{9}{*}{\centering GFCC} 
        & MobileNetV2 & 0.848 & 7.609
        \\ \cline{2-4}\cline{6-8}
        & ResNet50V2  & 0.776 & 1.873 && ResNet50V2  & 0.786 & 5.045  \\ \cline{2-4}\cline{6-8}
        & ResNet152V2 & 0.734 & 2.385 && ResNet152V2 & 0.864 & 1.679  \\ \cline{2-4}\cline{6-8}
        & VGG16       & 0.765 & 1.288 && VGG16       & 0.843 & 0.555  \\ \cline{2-4}\cline{6-8}
        & VGG19       & 0.734 & 0.498 && VGG19       & 0.828 & 0.496  \\ \cline{2-4}\cline{6-8}
        & DenseNet121 & 0.739 & 3.185 && DenseNet121 & 0.838 & 2.664  \\ \cline{2-4}\cline{6-8}
        & DenseNet201 & 0.651 & 2.771 && DenseNet201 & 0.817 & 2.318  \\ \cline{2-4}\cline{6-8}
        & Xception    & 0.760 & 8.531 && Xception    & 0.807 & 7.971  \\ \cline{2-4}\cline{6-8}
        & InceptionV3 & 0.765 & 3.113 && InceptionV3 & 0.833 & 4.667  \\ \hline
        \multirow{9}{*}{\centering CQCC} 
        & MobileNetV2 & 0.994 & 0.008 
        & \multirow{9}{*}{\centering BFCC} 
        & MobileNetV2 & 0.791 & 7.068  
        \\ \cline{2-4}\cline{6-8}
        & ResNet50V2  & 0.979 & 0.109 && ResNet50V2  & 0.796 & 4.732  \\ \cline{2-4}\cline{6-8}
        & ResNet152V2 & 0.979 & 0.161 && ResNet152V2 & 0.723 & 7.052  \\ \cline{2-4}\cline{6-8}
        & VGG16       & 0.994 & 0.017 && VGG16       & 0.786 & 0.470  \\ \cline{2-4}\cline{6-8}
        & VGG19       & 0.989 & 0.019 && VGG19       & 0.770 & 0.499  \\ \cline{2-4}\cline{6-8}
        & DenseNet121 & 1.000 & 0.000 && DenseNet121 & 0.812 & 2.268  \\ \cline{2-4}\cline{6-8}
        & DenseNet201 & 0.984 & 0.035 && DenseNet201 & 0.718 & 3.300  \\ \cline{2-4}\cline{6-8}
        & Xception    & 1.000 & 0.000 && Xception    & 0.770 & 9.317  \\ \cline{2-4}\cline{6-8}
        & InceptionV3 & 0.994 & 0.012 && InceptionV3 & 0.791 & 2.716  \\ \hline
        \multirow{9}{*}{\centering MFCC} 
        & MobileNetV2 & 0.817 & 7.428 
        & \multirow{9}{*}{\centering \shortstack{Mel\\Spectro}} 
        & MobileNetV2 & 0.838 & 9.244  
        \\ \cline{2-4}\cline{6-8}
        & ResNet50V2  & 0.848 & 1.549 && ResNet50V2  & 0.833 & 1.923  \\ \cline{2-4}\cline{6-8}
        & ResNet152V2 & 0.557 & 14.10 && ResNet152V2 & 0.843 & 3.693  \\ \cline{2-4}\cline{6-8}
        & VGG16       & 0.791 & 0.498 && VGG16       & 0.765 & 0.396  \\ \cline{2-4}\cline{6-8}
        & VGG19       & 0.520 & 0.692 && VGG19       & 0.875 & 0.315  \\ \cline{2-4}\cline{6-8}
        & DenseNet121 & 0.869 & 1.374 && DenseNet121 & 0.791 & 3.158  \\ \cline{2-4}\cline{6-8}
        & DenseNet201 & 0.807 & 4.450 && DenseNet201 & 0.859 & 2.950  \\ \cline{2-4}\cline{6-8}
        & Xception    & 0.864 & 3.286 && Xception    & 0.828 & 4.113  \\ \cline{2-4}\cline{6-8}
        & InceptionV3 & 0.880 & 1.264 && InceptionV3 & 0.802 & 1.753  \\ \hline
        \multirow{9}{*}{\centering Chroma} 
        & MobileNetV2 & 0.781 & 7.462 
        & \multirow{9}{*}{\centering \shortstack{Spectral\\Centroid}} 
        & MobileNetV2 & 0.854 & 8.867  
        \\ \cline{2-4}\cline{6-8}
        & ResNet50V2  & 0.750 & 2.337 && ResNet50V2  & 0.854 & 2.005  \\ \cline{2-4}\cline{6-8}
        & ResNet152V2 & 0.781 & 3.758 && ResNet152V2 & 0.822 & 1.076  \\ \cline{2-4}\cline{6-8}
        & VGG16       & 0.776 & 0.777 && VGG16       & 0.859 & 0.273  \\ \cline{2-4}\cline{6-8}
        & VGG19       & 0.677 & 0.620 && VGG19       & 0.817 & 0.426  \\ \cline{2-4}\cline{6-8}
        & DenseNet121 & 0.734 & 3.294 && DenseNet121 & 0.869 & 1.361  \\ \cline{2-4}\cline{6-8}
        & DenseNet201 & 0.791 & 5.286 && DenseNet201 & 0.869 & 1.479  \\ \cline{2-4}\cline{6-8}
        & Xception    & 0.817 & 4.872 && Xception    & 0.875 & 2.733  \\ \cline{2-4}\cline{6-8}
        & InceptionV3 & 0.697 & 3.928 && InceptionV3 & 0.859 & 3.972  \\ \hline
    \end{tabular}
    \egroup
    \label{tb:single_feat_results}
\end{table}

\begin{table}[H]
    \scriptsize
    \centering
    \caption{Combined features experimental results.}
    \bgroup
    \def\arraystretch{1.5}%
    \begin{tabular}{|l|l|l|l|l|l|l|l|} 
        \hline
        \textbf{Features} & \textbf{Model} & \textbf{Accuracy} & \textbf{Loss} & \textbf{Features} & \textbf{Model} & \textbf{Accuracy} & \textbf{Loss}  \\ \hline
        \multirow{9}{*}{\centering \shortstack{CQCC \\\\\\\\ MFCC \\\\\\\\ BFCC}}
        & MobileNetV2 & \textbf{1.000} & \textbf{0.000} 
        & \multirow{9}{*}{\centering \shortstack{GFCC \\\\\\\\ BFCC \\\\\\\\ CQCC}}
        & MobileNetV2 & 0.890 & 9.393  
        \\ \cline{2-4}\cline{6-8}
        & ResNet50V2  & \textbf{1.000} & \textbf{0.000} && ResNet50V2  & 0.854 & 3.961  \\ \cline{2-4}\cline{6-8}
        & ResNet152V2 & \textbf{1.000} & \textbf{0.000} && ResNet152V2 & 0.885 & 9.836  \\ \cline{2-4}\cline{6-8}
        & VGG16       & \textbf{1.000} & \textbf{0.000} && VGG16       & 0.854 & 0.355  \\ \cline{2-4}\cline{6-8}
        & VGG19       & \textbf{1.000} & \textbf{0.000} && VGG19       & 0.760 & 0.481  \\ \cline{2-4}\cline{6-8}
        & DenseNet121 & \textbf{1.000} & \textbf{0.000} && DenseNet121 & 0.828 & 1.048  \\ \cline{2-4}\cline{6-8}
        & DenseNet201 & \textbf{1.000} & \textbf{0.000} && DenseNet201 & 0.859 & 1.594  \\ \cline{2-4}\cline{6-8}
        & Xception    & \textbf{1.000} & \textbf{0.000} && Xception    & 0.848 & 9.100  \\ \cline{2-4}\cline{6-8}
        & InceptionV3 & \textbf{1.000} & \textbf{0.000} && InceptionV3 & 0.781 & 4.304  \\ \hline
        \multirow{9}{*}{\centering \shortstack{MFCC \\\\\\\\ Chroma \\\\\\\\ Mel\\Spectro}}
        & MobileNetV2 & 0.875 & 6.764 
        & \multirow{9}{*}{\centering \shortstack{GFCC \\\\\\\\ CQCC \\\\\\\\ LFCC}}
        & MobileNetV2 & 0.979 & 0.487  
        \\ \cline{2-4}\cline{6-8}
        & ResNet50V2  & 0.786 & 9.704 && ResNet50V2  & 0.838 & 3.509  \\ \cline{2-4}\cline{6-8}
        & ResNet152V2 & 0.875 & 2.720 && ResNet152V2 & 0.979 & 0.141  \\ \cline{2-4}\cline{6-8}
        & VGG16       & 0.833 & 0.388 && VGG16       & 0.786 & 3.453  \\ \cline{2-4}\cline{6-8}
        & VGG19       & 0.817 & 0.548 && VGG19       & 0.708 & 0.475  \\ \cline{2-4}\cline{6-8}
        & DenseNet121 & 0.838 & 1.894 && DenseNet121 & 0.963 & 0.294  \\ \cline{2-4}\cline{6-8}
        & DenseNet201 & 0.833 & 1.945 && DenseNet201 & 0.989 & 0.229  \\ \cline{2-4}\cline{6-8}
        & Xception    & 0.890 & 2.618 && Xception    & 0.864 & 5.917  \\ \cline{2-4}\cline{6-8}
        & InceptionV3 & 0.796 & 5.233 && InceptionV3 & 0.828 & 2.079  \\ \hline
        \multirow{9}{*}{\centering \shortstack{MFCC \\\\\\\\ Chroma \\\\\\\\ LFCC}}
        & MobileNetV2 & 0.796 & 0.764 
        & \multirow{9}{*}{\centering \shortstack{Mel\\Spectro \\\\\\\\ MFCC \\\\\\\\ BFCC}}
        & MobileNetV2 & 0.864 & 6.603  
        \\ \cline{2-4}\cline{6-8}
        & ResNet50V2  & 0.807 & 3.938 && ResNet50V2  & 0.833 & 2.246  \\ \cline{2-4}\cline{6-8}
        & ResNet152V2 & 0.807 & 1.653 && ResNet152V2 & 0.869 & 1.940  \\ \cline{2-4}\cline{6-8}
        & VGG16       & 0.708 & 0.541 && VGG16       & 0.828 & 0.461  \\ \cline{2-4}\cline{6-8}
        & VGG19       & 0.817 & 1.222 && VGG19       & 0.765 & 0.554  \\ \cline{2-4}\cline{6-8}
        & DenseNet121 & 0.812 & 1.969 && DenseNet121 & 0.848 & 2.250  \\ \cline{2-4}\cline{6-8}
        & DenseNet201 & 0.807 & 4.927 && DenseNet201 & 0.822 & 2.373  \\ \cline{2-4}\cline{6-8}
        & Xception    & 0.796 & 4.091 && Xception    & 0.838 & 5.584  \\ \cline{2-4}\cline{6-8}
        & InceptionV3 & 0.833 & 3.257 && InceptionV3 & 0.807 & 1.684  \\ \hline
    \end{tabular}
    \egroup
    \label{tb:multi_feat_results}
\end{table}

\begin{figure}[h]
\centering
\includegraphics[width=\textwidth]{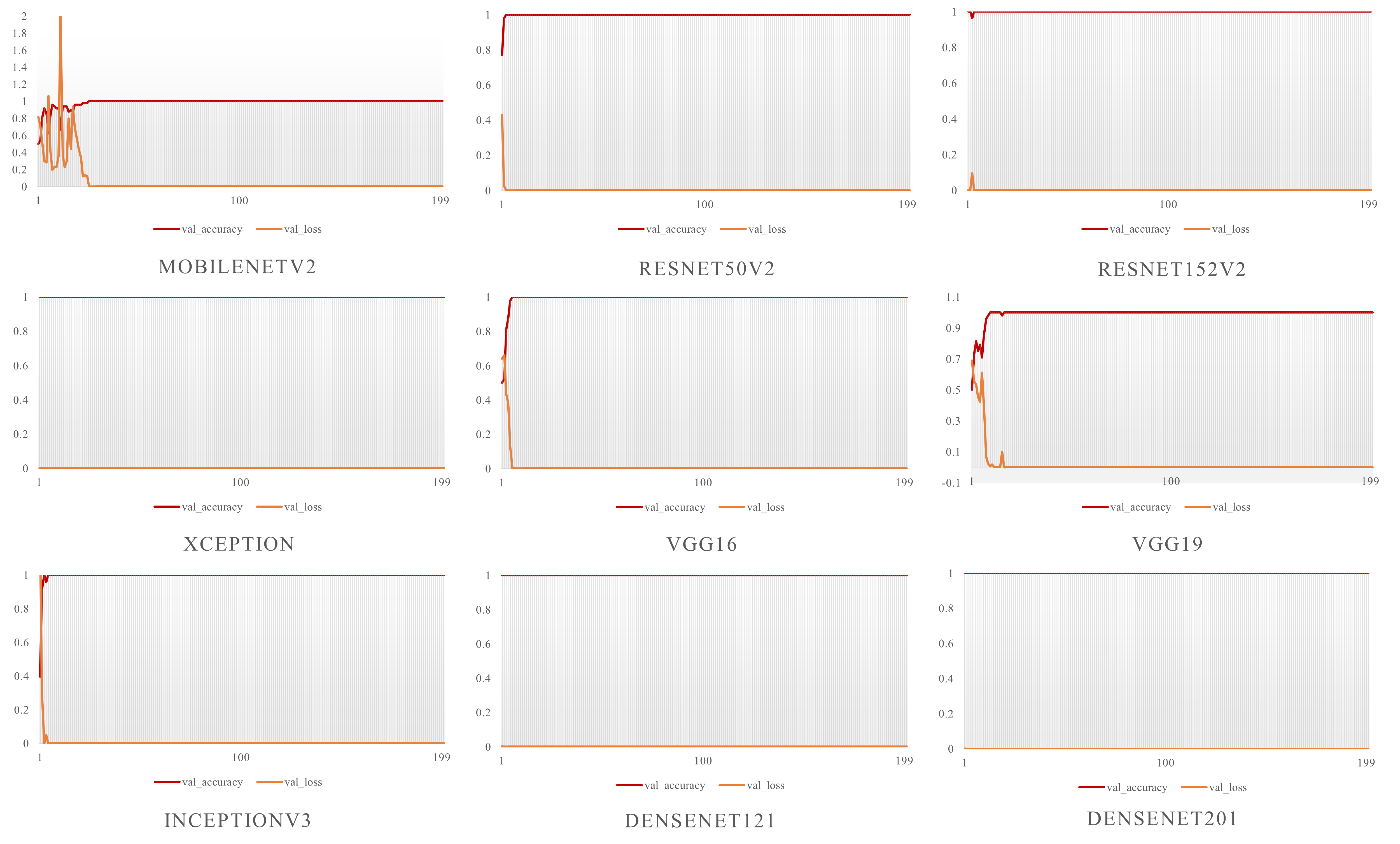}
\caption{Validation accuracy and loss progress of CQCC-MFCC-BFCC on RPW dataset.} 
\label{fig:progress-200}
\end{figure}

Figure \ref{fig:progress-200} shows the validation accuracy and loss progress of all the used DL models during the training; each line consists of 200 points, one for each epoch. 

\begin{figure}[h]
\centering
\includegraphics[width=\textwidth]{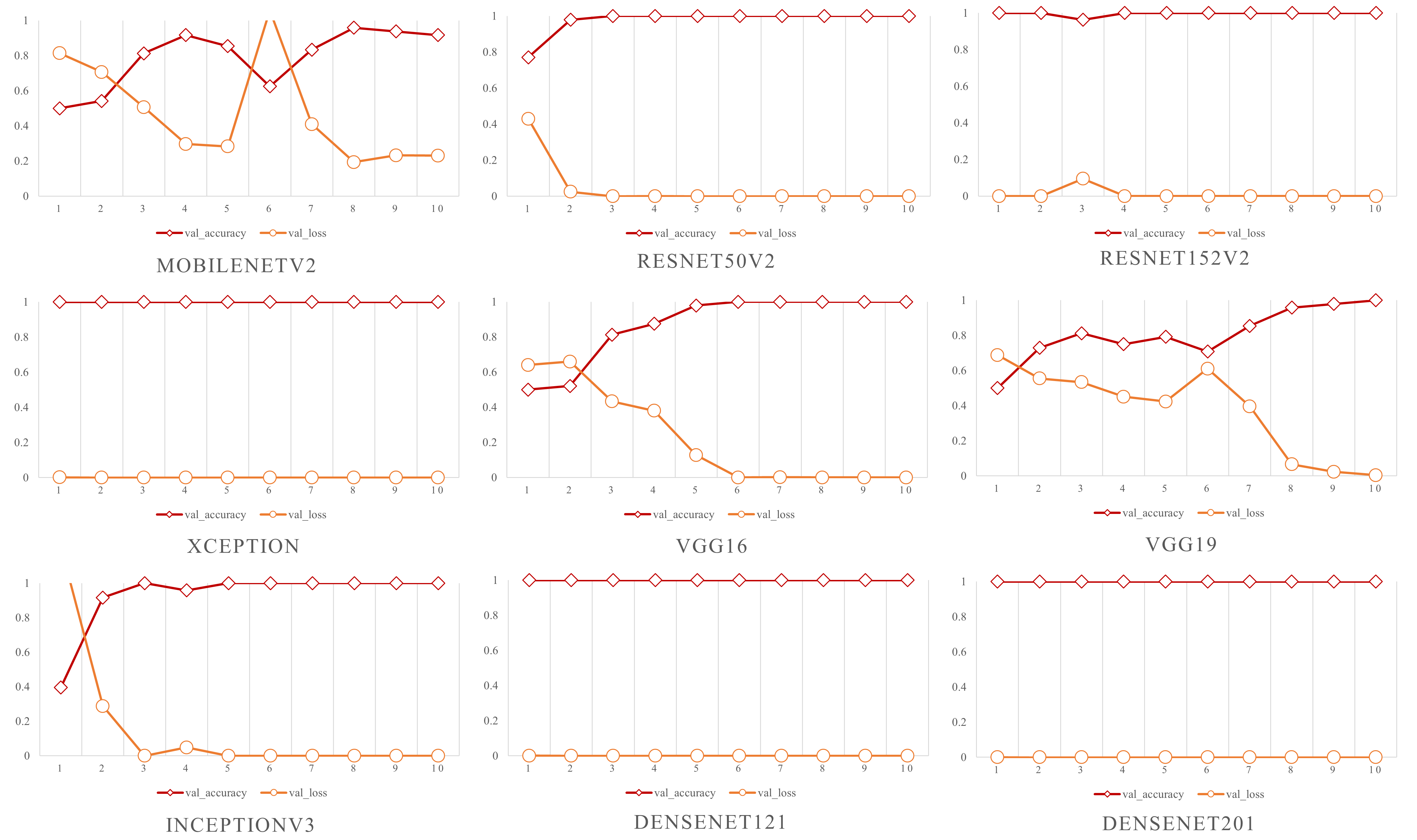}
\caption{Validation accuracy and loss progress of CQCC-MFCC-BFCC on RPW dataset for the first 10 epochs.} 
\label{fig:progress-10}
\end{figure}

Figure \ref{fig:progress-10} presents a clearer version of Figure \ref{fig:progress-200}, showing how fast the models are learning during the first 10 epochs.

Based on figures \ref{fig:progress-200} and \ref{fig:progress-10}, it appears that all of the models were able to reach a validation accuracy of 100\% within the first 10 epochs when using the merged features. The full distribution plot provides a more comprehensive view of the models' performance throughout the training. In contrast, the plot of the first 10 epochs provides a more focused view of the models' initial performance. This confirms that all models were able to effectively learn and make accurate predictions on the validation set within a relatively short period of time. One possible explanation for this is that the models were pre-trained on a large dataset, which allowed them to effectively learn the features of the validation set in a relatively short time. Additionally, the high accuracy achieved within the first 10 epochs also suggests that the models were able to effectively generalize to unseen data, which is an important characteristic of a good model.

\begin{figure}[!htp]
\centering
\includegraphics[width=\textwidth]{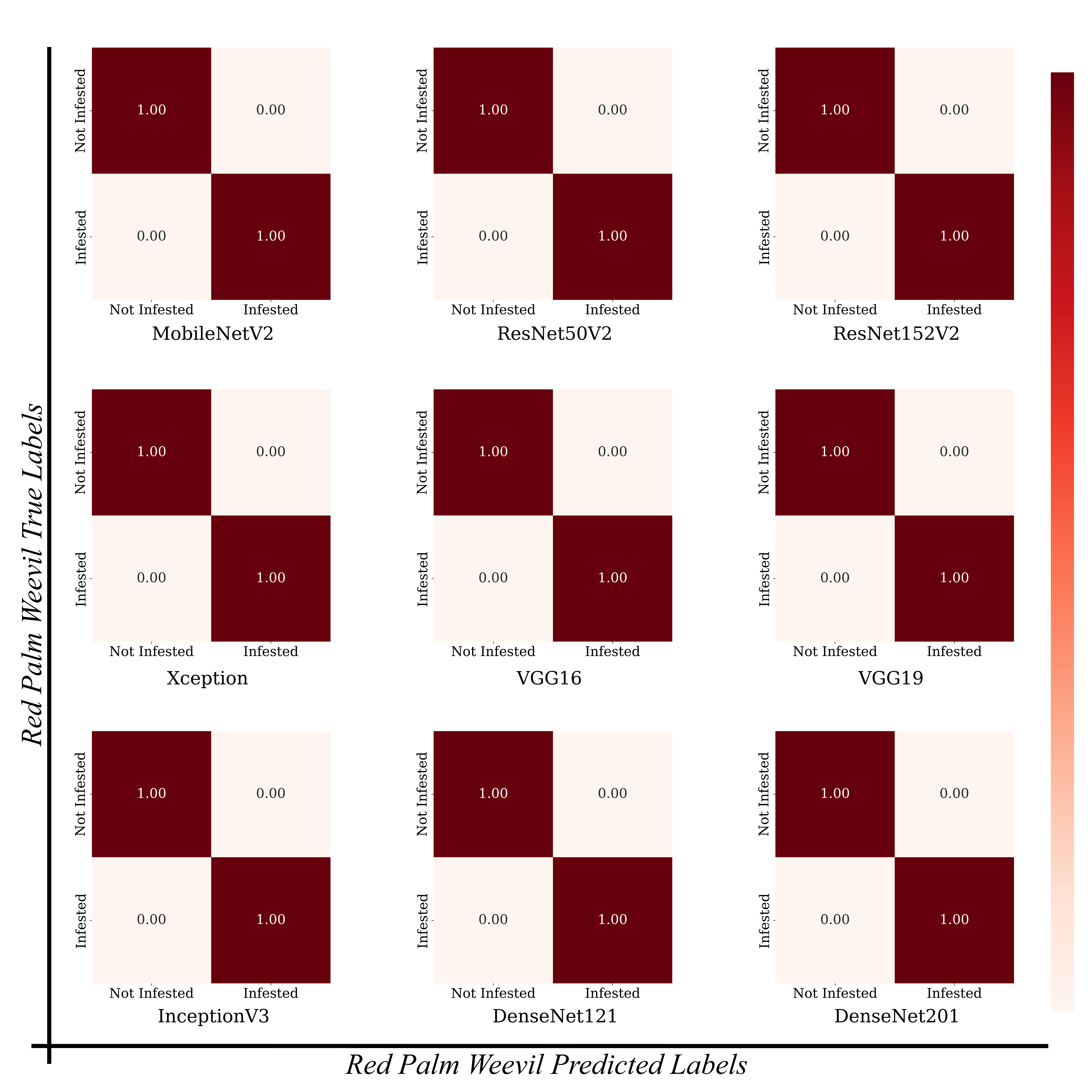}
\caption{Confusion matrices of CQCC-MFCC-BFCC on RPW dataset.}
\label{fig:cm}
\end{figure}

The confusion matrix results presented in Figure \ref{fig:cm}  are impressive, as they show that all of the provided models were able to achieve 100\% accuracy on the test set. This suggests that the models have been effectively trained and are able to make accurate predictions on unseen data, which is a key indicator of a high-performing model.

\section{Evaluation}

In this section, several experiments have been conducted to evaluate the performances of the proposed approach using three public datasets. The first dataset, TreeVibes\footnote{\url {https://www.kaggle.com/datasets/potamitis/treevibes}}, contains recordings of internal vibrations of trees that are wirelessly transmitted to a cloud server using the device detailed in Section 3. The dataset aims to detect the presence of wood-boring insects that feed or move inside trees, which can cause high tree mortality. The dataset contains almost 1000 "clean" label recordings and over 50000 "infested" label recordings, making it a valuable resource for detecting tree infestations and preventing damage caused by insect feeding.

We applied the InceptionV3 model with various features, including CQCC, MFCC, BFCC, Spectral-Centroid \cite{ESMAILKARAR20225309} and the proposed combined features (CQCC-MFCC-BFCC). The results showed a 100\% test accuracy in all the cases, indicating that the dataset may be too simple for DL methods. Since this evaluation section aims to assess the effectiveness of combining features in improving the accuracy of detecting tree infestations, we will use machine learning techniques algorithms such as Support Vector Machine (SVM), Logistic Regression (LR), Decision Trees (DT), and Random Forest (RF). The results showed that using different features, the accuracy ranged from 0.980 to 1.000, with the best accuracy achieved using the proposed combined features with the LR method. Table \ref{tab:treevibes} shows the detailed results of the experiments on the TreeVibes dataset. We note that the combination of the three features (CQCC-MFCC-BFCC) provides the best accuracy of detecting tree infestations for all four machine learning techniques.

\begin{table}[h]
\scriptsize
\centering
\begin{tabular}{|c|c|c|c|c|}
\hline
\textbf{Features} & \textbf{SVM} & \textbf{LR} & \textbf{DT} & \textbf{RF}  \\ \hline
CQCC & 0.980 & 0.992 & 0.980 & 0.990  \\ \hline
MFCC & 0.982 & 0.998 & 0.990 & 0.997  \\ \hline
BFCC & 0.982 & 0.992 & 0.980 & 0.986  \\ \hline
Karar et al. \cite{ESMAILKARAR20225309} & 0.995 & 0.995 & 0.995 & 0.995  \\ \hline
\textbf{Proposed approach} & \textbf{0.997} & \textbf{1.000} & \textbf{0.995} & \textbf{0.997}  \\ \hline
\end{tabular}
 \caption{Experimental results on Treevibes dataset. \label{tab:treevibes}}
\end{table}

To better validate the performance of the proposed approach, we used two public datasets related to ESC, ESC-10\footnote{\url{https://github.com/karolpiczak/ESC-10}} and ESC-50\footnote{\url{https://github.com/karolpiczak/ESC-50}}. The two datasets consist of several brief environment recordings. Each recording consists of five seconds. We split the ESC-10 and ESC-50 datasets into train:val:test splits with ratios of 0.8:0.1:0.1.

For the ESC-10 dataset, we trained the InceptionV3 model on five different feature extraction techniques, including CQCC, MFCC, BFCC, LBP-VAR-ELBP-LBP1D-LPQ1D-MFCC-GFCC \cite{toffa2020environmental}, and our proposed combination of CQCC-MFCC-BFCC. Results showed that the proposed combination achieved the highest validation accuracy of 0.95 with a precision of 0.947, recall of 1.0, and an F1 score of 0.973. The second-highest accuracy of 0.925 was achieved by the LBP-VAR-ELBP-LBP1D-LPQ1D-MFCC-GFCC combination. The accuracies of the models trained on CQCC, MFCC, and BFCC were 0.875, 0.875, and 0.825, respectively. Table \ref{tab:esc10} shows the detailed results of the experiments on ESC-10 dataset.

Similarly, for the ESC-50 dataset, we trained the InceptionV3 model using the same five feature extraction techniques. The proposed combination of CQCC-MFCC-BFCC achieved the highest accuracy of 0.995 with a precision of 0.995, a recall of 1.0, and an F1 score of 0.997. The InceptionV3 model trained using the LBP-VAR-ELBP-LBP1D-LPQ1D-MFCC-GFCC \cite{toffa2020environmental} combination achieved the second-highest accuracy of 0.985, with a precision of 0.995, recall of 0.99, and an F1 score of 0.992. The models trained using CQCC, MFCC, and BFCC achieved accuracies of 0.975, 0.96, and 0.98, respectively. Table \ref{tab:esc50} shows the detailed results of the experiments on the ESC-50 dataset. Based on these results, we can conclude that the proposed combination of CQCC-MFCC-BFCC achieved the highest accuracy for both ESC-10 and ESC-50 datasets.

The InceptionV3 model trained using the LBP-VAR-ELBP-LBP1D-LPQ1D-MFCC-GFCC combination \cite{toffa2020environmental} also performed well for both datasets, achieving the second-highest accuracy. Moreover, the InceptionV3 model is a fast model, making it a good choice for these datasets. It is important to note that the choice of feature extraction technique could impact the final accuracy results, and the learning rate used during training could also affect the performance of the model.

\begin{table}[h]
\scriptsize
\centering
\begin{tabular}{|c|c|c|c|c|}
\hline
\textbf{Features} & \textbf{Accuracy} & \textbf{Precision} & \textbf{Recall} & \textbf{F1 Score}  \\ \hline
CQCC & 0.875 & 0.897 & 0.972 & 0.933  \\ \hline
MFCC & 0.875 & 0.919 & 0.944 & 0.932  \\ \hline
BFCC & 0.825 & 0.892 & 0.917 & 0.904  \\ \hline
Toffa et al. \cite{toffa2020environmental} & 0.925 & \textbf{0.971} & 0.944 & 0.958  \\ \hline
\textbf{Proposed approach} &  \textbf{0.950} & 0.947 & \textbf{1.000} & \textbf{0.973} \\ \hline
\end{tabular}
 \caption{Experimental results on the ESC10 dataset. \label{tab:esc10}}
\end{table}

\begin{table}[h]
\scriptsize
\centering
\begin{tabular}{|c|c|c|c|c|}
\hline
\textbf{Features} & \textbf{Accuracy} & \textbf{Precision} & \textbf{Recall} & \textbf{F1 Score}  \\ \hline
CQCC & 0.975 & 0.995 & 0.980 & 0.987  \\ \hline
MFCC & 0.960 & 0.990 & 0.969 & 0.979  \\ \hline
BFCC & 0.980 & 0.990 & 0.990 & 0.990  \\ \hline
Toffa et al. \cite{toffa2020environmental} & 0.985 & 0.995 & 0.990 & 0.992  \\ \hline
\textbf{Proposed approach} & \textbf{0.995} & \textbf{0.995} & \textbf{1.000} & \textbf{0.997}  \\ \hline
\end{tabular}
 \caption{Experimental results on ESC50 dataset. \label{tab:esc50}}
\end{table}

Acoustic-based detection systems can complement image-based techniques for detecting plant diseases and infestations, especially when external symptoms are not apparent. In many situations, sound can be affected by background noise. Hence, DL techniques constitute an excellent solution since they are very effective in detecting RPW, even in background noise. This is because they can learn to distinguish between the characteristics of the RPW's acoustic signature and the noise. The acoustic signals of the RPW have a unique frequency range, waveform pattern, and duration that differ from those of other sounds that occur in the environment. DL models, such as convolutional neural networks (CNNs), can be trained to recognize these unique characteristics and distinguish them from background noise. However, it is essential to note that certain limitations need to be addressed to implement the proposed solution in a real-world scenario. First, we acknowledge that the TreeVibe device incurs a high cost for acquisition and maintenance, which may pose a challenge for widespread adoption. Additionally, the considered device is susceptible to noise if not correctly fitted or calibrated, which can impact the accuracy and reliability of the detection results. The proposed solution can be integrated into an architecture based on Optical Fiber Distributed Acoustic Sensing, as proposed in the paper \cite{s22176491}. Identifying the locations where the devices will be deployed is crucial, and this will involve mapping the farm and identifying the areas where RPW infestations are most prevalent.

\section{Conclusions}
Early detection of RPW infestation can play a crucial role in reducing the negative impact of this pest on palm trees and the industries that rely on them. Early detection can help to reduce the spread of the infestation, as prompt action can be taken to contain and control the population. This can prevent the red weevil from spreading to other trees, which can help to preserve the overall health of the palm population. Additionally, early detection can also help to reduce the cost of control measures. By taking action before significant damage occurs, less extensive and costly control methods may be required. This can also help to reduce the overall economic impact of red weevil infestations on the palm industry. This study suggests using a DL approach to detect RPW infestation in palm tree farms. The proposed approach utilizes three features, namely CQCC, MFCC, and BFCC, to convert sound data received from sensors embedded in palm trees into images. The classification of resulting images into two classes (infested and not infested), based on several DL techniques, shows good performances of the proposed approach on a dataset representing the Al-Ahssa region of the Kingdom of Saudi Arabia, as well as public datasets. The proposed approach can help to improve the accuracy and efficiency of detecting and managing RPW. This study can be extended considering several perspectives. First, the proposed system can be integrated with existing farm management tools to enable automated pest control and monitoring. In addition, an important research area to explore will be the investigation of other advanced DL techniques, such as reinforcement learning or generative models, to improve the performance of the detection system and to predict the severity of the infestation and the potential damage to the tree. Finally, this paper can be extended by conducting more field trials in various settings. This could involve testing the system's effectiveness in different geographical locations, under varying climatic conditions, and with different species of palm trees, which could vary in their acoustic properties. Additionally, the impact of various background noises that might be present in recordings can be further explored. Such extensive field trials would provide a more robust understanding of the system's performance across diverse real-world conditions.

\vspace{6pt}

\section*{Acknowledgement}
The authors would like to express their deepest gratitude to Prof. Abdulrahman S. Aldawood for providing the dataset. Also, the authors would like to thank Prince Sultan University for their support.

\bibliographystyle{elsarticle-harv}

\bibliography{references.bib}

\begin{thebibliography}{50}
\expandafter\ifx\csname natexlab\endcsname\relax\def\natexlab#1{#1}\fi
\providecommand{\url}[1]{\texttt{#1}}
\providecommand{\href}[2]{#2}
\providecommand{\path}[1]{#1}
\providecommand{\DOIprefix}{}
\providecommand{\ArXivprefix}{arXiv:}
\providecommand{\URLprefix}{}
\providecommand{\Pubmedprefix}{pmid:}
\providecommand{\doi}[1]{\href{http://dx.doi.org/#1}{\path{#1}}}
\providecommand{\Pubmed}[1]{\href{pmid:#1}{\path{#1}}}
\providecommand{\bibinfo}[2]{#2}
\ifx\xfnm\relax \def\xfnm[#1]{\unskip,\space#1}\fi
\bibitem[{Abbaskhah et~al.(2023)Abbaskhah, Sedighi and
  Marvi}]{ABBASKHAH2023105261}
\bibinfo{author}{Abbaskhah, A.}, \bibinfo{author}{Sedighi, H.},
  \bibinfo{author}{Marvi, H.}, \bibinfo{year}{2023}.
\newblock \bibinfo{title}{Infant cry classification by mfcc feature extraction
  with mlp and cnn structures}.
\newblock \bibinfo{journal}{Biomedical Signal Processing and Control}
  \bibinfo{volume}{86}, \bibinfo{pages}{105261}.
\newblock \URLprefix
  \url{https://www.sciencedirect.com/science/article/pii/S1746809423006948},
  \DOIprefix\doi{https://doi.org/10.1016/j.bspc.2023.105261}.
\bibitem[{Al-Sarem et~al.(2021)Al-Sarem, Alsaeedi, Saeed, Boulila and
  AmeerBakhsh}]{al2021novel}
\bibinfo{author}{Al-Sarem, M.}, \bibinfo{author}{Alsaeedi, A.},
  \bibinfo{author}{Saeed, F.}, \bibinfo{author}{Boulila, W.},
  \bibinfo{author}{AmeerBakhsh, O.}, \bibinfo{year}{2021}.
\newblock \bibinfo{title}{A novel hybrid deep learning model for detecting
  covid-19-related rumors on social media based on lstm and concatenated
  parallel cnns}.
\newblock \bibinfo{journal}{Applied Sciences} \bibinfo{volume}{11},
  \bibinfo{pages}{7940}.
\bibitem[{Allen(1977)}]{allen1977short}
\bibinfo{author}{Allen, J.}, \bibinfo{year}{1977}.
\newblock \bibinfo{title}{Short term spectral analysis, synthesis, and
  modification by discrete fourier transform}.
\newblock \bibinfo{journal}{IEEE Transactions on Acoustics, Speech, and Signal
  Processing} \bibinfo{volume}{25}, \bibinfo{pages}{235--238}.
\bibitem[{Alom et~al.(2018)Alom, Taha, Yakopcic, Westberg, Sidike, Nasrin,
  Van~Esesn, Awwal and Asari}]{alom2018history}
\bibinfo{author}{Alom, M.Z.}, \bibinfo{author}{Taha, T.M.},
  \bibinfo{author}{Yakopcic, C.}, \bibinfo{author}{Westberg, S.},
  \bibinfo{author}{Sidike, P.}, \bibinfo{author}{Nasrin, M.S.},
  \bibinfo{author}{Van~Esesn, B.C.}, \bibinfo{author}{Awwal, A.A.S.},
  \bibinfo{author}{Asari, V.K.}, \bibinfo{year}{2018}.
\newblock \bibinfo{title}{The history began from alexnet: A comprehensive
  survey on deep learning approaches}.
\newblock \bibinfo{journal}{arXiv preprint arXiv:1803.01164} .
\bibitem[{Ashry et~al.(2022)Ashry, Wang, Mao, Sait, Guo, Al-Fehaid, Al-Shawaf,
  Ng and Ooi}]{s22176491}
\bibinfo{author}{Ashry, I.}, \bibinfo{author}{Wang, B.}, \bibinfo{author}{Mao,
  Y.}, \bibinfo{author}{Sait, M.}, \bibinfo{author}{Guo, Y.},
  \bibinfo{author}{Al-Fehaid, Y.}, \bibinfo{author}{Al-Shawaf, A.},
  \bibinfo{author}{Ng, T.K.}, \bibinfo{author}{Ooi, B.S.},
  \bibinfo{year}{2022}.
\newblock \bibinfo{title}{Cnn-aided optical fiber distributed acoustic sensing
  for early detection of red palm weevil: A field experiment}.
\newblock \bibinfo{journal}{Sensors} \bibinfo{volume}{22}.
\newblock \DOIprefix\doi{10.3390/s22176491}.
\bibitem[{Ben~Atitallah et~al.(2022a)Ben~Atitallah, Driss, Boulila and
  Ben~Ghezala}]{ben2022randomly}
\bibinfo{author}{Ben~Atitallah, S.}, \bibinfo{author}{Driss, M.},
  \bibinfo{author}{Boulila, W.}, \bibinfo{author}{Ben~Ghezala, H.},
  \bibinfo{year}{2022}a.
\newblock \bibinfo{title}{Randomly initialized convolutional neural network for
  the recognition of covid-19 using x-ray images}.
\newblock \bibinfo{journal}{International journal of imaging systems and
  technology} \bibinfo{volume}{32}, \bibinfo{pages}{55--73}.
\bibitem[{Ben~Atitallah et~al.(2022b)Ben~Atitallah, Driss, Boulila, Koubaa and
  Ben~Ghezala}]{ben2022fusion}
\bibinfo{author}{Ben~Atitallah, S.}, \bibinfo{author}{Driss, M.},
  \bibinfo{author}{Boulila, W.}, \bibinfo{author}{Koubaa, A.},
  \bibinfo{author}{Ben~Ghezala, H.}, \bibinfo{year}{2022}b.
\newblock \bibinfo{title}{Fusion of convolutional neural networks based on
  dempster--shafer theory for automatic pneumonia detection from chest x-ray
  images}.
\newblock \bibinfo{journal}{International Journal of Imaging Systems and
  Technology} \bibinfo{volume}{32}, \bibinfo{pages}{658--672}.
\bibitem[{Boulila et~al.(2022)Boulila, Khlifi, Ammar, Koubaa, Benjdira and
  Farah}]{boulila2022hybrid}
\bibinfo{author}{Boulila, W.}, \bibinfo{author}{Khlifi, M.K.},
  \bibinfo{author}{Ammar, A.}, \bibinfo{author}{Koubaa, A.},
  \bibinfo{author}{Benjdira, B.}, \bibinfo{author}{Farah, I.R.},
  \bibinfo{year}{2022}.
\newblock \bibinfo{title}{A hybrid privacy-preserving deep learning approach
  for object classification in very high-resolution satellite images}.
\newblock \bibinfo{journal}{Remote Sensing} \bibinfo{volume}{14},
  \bibinfo{pages}{4631}.
\bibitem[{Brown(1991)}]{brown1991calculation}
\bibinfo{author}{Brown, J.C.}, \bibinfo{year}{1991}.
\newblock \bibinfo{title}{Calculation of a constant q spectral transform}.
\newblock \bibinfo{journal}{The Journal of the Acoustical Society of America}
  \bibinfo{volume}{89}, \bibinfo{pages}{425--434}.
\bibitem[{Bu and Wang(2019)}]{bu2019smart}
\bibinfo{author}{Bu, F.}, \bibinfo{author}{Wang, X.}, \bibinfo{year}{2019}.
\newblock \bibinfo{title}{A smart agriculture iot system based on deep
  reinforcement learning}.
\newblock \bibinfo{journal}{Future Generation Computer Systems}
  \bibinfo{volume}{99}, \bibinfo{pages}{500--507}.
\bibitem[{Chen et~al.(2021)Chen, Sun, Chen, Xie, Wu and Xu}]{chen2021deep}
\bibinfo{author}{Chen, W.}, \bibinfo{author}{Sun, Q.}, \bibinfo{author}{Chen,
  X.}, \bibinfo{author}{Xie, G.}, \bibinfo{author}{Wu, H.},
  \bibinfo{author}{Xu, C.}, \bibinfo{year}{2021}.
\newblock \bibinfo{title}{Deep learning methods for heart sounds
  classification: a systematic review}.
\newblock \bibinfo{journal}{Entropy} \bibinfo{volume}{23},
  \bibinfo{pages}{667}.
\bibitem[{Chollet(2017)}]{xception}
\bibinfo{author}{Chollet, F.}, \bibinfo{year}{2017}.
\newblock \bibinfo{title}{Xception: Deep learning with depthwise separable
  convolutions}.
\newblock \href{http://arxiv.org/abs/1610.02357v3}{{\tt arXiv:1610.02357v3}}.
\bibitem[{El-Juhany et~al.(2010)}]{el2010degradation}
\bibinfo{author}{El-Juhany, L.I.}, et~al., \bibinfo{year}{2010}.
\newblock \bibinfo{title}{Degradation of date palm trees and date production in
  arab countries: causes and potential rehabilitation}.
\newblock \bibinfo{journal}{Australian Journal of Basic and Applied Sciences}
  \bibinfo{volume}{4}, \bibinfo{pages}{3998--4010}.
\bibitem[{{Esmail Karar} et~al.(2022){Esmail Karar}, Abdel-Aty, Algarni,
  {Fadzil Hassan}, Abdou and Reyad}]{ESMAILKARAR20225309}
\bibinfo{author}{{Esmail Karar}, M.}, \bibinfo{author}{Abdel-Aty, A.H.},
  \bibinfo{author}{Algarni, F.}, \bibinfo{author}{{Fadzil Hassan}, M.},
  \bibinfo{author}{Abdou, M.}, \bibinfo{author}{Reyad, O.},
  \bibinfo{year}{2022}.
\newblock \bibinfo{title}{Smart iot-based system for detecting rpw larvae in
  date palms using mixed depthwise convolutional networks}.
\newblock \bibinfo{journal}{Alexandria Engineering Journal}
  \bibinfo{volume}{61}, \bibinfo{pages}{5309--5319}.
\bibitem[{Ferreira et~al.(2021)Ferreira, Lotte, D'Elia, Stamatopoulos, Kim and
  Benjamin}]{ferreira2021accurate}
\bibinfo{author}{Ferreira, M.P.}, \bibinfo{author}{Lotte, R.G.},
  \bibinfo{author}{D'Elia, F.V.}, \bibinfo{author}{Stamatopoulos, C.},
  \bibinfo{author}{Kim, D.H.}, \bibinfo{author}{Benjamin, A.R.},
  \bibinfo{year}{2021}.
\newblock \bibinfo{title}{Accurate mapping of brazil nut trees (bertholletia
  excelsa) in amazonian forests using worldview-3 satellite images and
  convolutional neural networks}.
\newblock \bibinfo{journal}{Ecological Informatics} \bibinfo{volume}{63},
  \bibinfo{pages}{101302}.
\bibitem[{Gambhir et~al.(2023)Gambhir, Dev, Bansal and Sharma}]{gambhir2023end}
\bibinfo{author}{Gambhir, P.}, \bibinfo{author}{Dev, A.},
  \bibinfo{author}{Bansal, P.}, \bibinfo{author}{Sharma, D.K.},
  \bibinfo{year}{2023}.
\newblock \bibinfo{title}{End-to-end multi-modal low-resourced speech keywords
  recognition using sequential conv2d nets}.
\newblock \bibinfo{journal}{ACM Transactions on Asian and Low-Resource Language
  Information Processing} .
\bibitem[{Ghandorh et~al.(2022)Ghandorh, Boulila, Masood, Koubaa, Ahmed and
  Ahmad}]{ghandorh2022semantic}
\bibinfo{author}{Ghandorh, H.}, \bibinfo{author}{Boulila, W.},
  \bibinfo{author}{Masood, S.}, \bibinfo{author}{Koubaa, A.},
  \bibinfo{author}{Ahmed, F.}, \bibinfo{author}{Ahmad, J.},
  \bibinfo{year}{2022}.
\newblock \bibinfo{title}{Semantic segmentation and edge detection—approach
  to road detection in very high resolution satellite images}.
\newblock \bibinfo{journal}{Remote Sensing} \bibinfo{volume}{14},
  \bibinfo{pages}{613}.
\bibitem[{Goldshtein et~al.(2022)Goldshtein, Soroker, Mandelik, Sadeh,
  Haberman, Ezra and Cohen}]{goldshtein2022analyzing}
\bibinfo{author}{Goldshtein, E.}, \bibinfo{author}{Soroker, V.},
  \bibinfo{author}{Mandelik, Y.}, \bibinfo{author}{Sadeh, A.},
  \bibinfo{author}{Haberman, A.}, \bibinfo{author}{Ezra, N.},
  \bibinfo{author}{Cohen, Y.}, \bibinfo{year}{2022}.
\newblock \bibinfo{title}{Analyzing spatiotemporal species spread by three
  declustering methods utilizing monitoring data based on national programs and
  citizen science}.
\newblock \bibinfo{journal}{Ecological Informatics} \bibinfo{volume}{72},
  \bibinfo{pages}{101916}.
\bibitem[{Haridasan et~al.(2023)Haridasan, Thomas and Raj}]{haridasan2023deep}
\bibinfo{author}{Haridasan, A.}, \bibinfo{author}{Thomas, J.},
  \bibinfo{author}{Raj, E.D.}, \bibinfo{year}{2023}.
\newblock \bibinfo{title}{Deep learning system for paddy plant disease
  detection and classification}.
\newblock \bibinfo{journal}{Environmental Monitoring and Assessment}
  \bibinfo{volume}{195}, \bibinfo{pages}{120}.
\bibitem[{He et~al.(2016a)He, Zhang, Ren and Sun}]{resnet50v2}
\bibinfo{author}{He, K.}, \bibinfo{author}{Zhang, X.}, \bibinfo{author}{Ren,
  S.}, \bibinfo{author}{Sun, J.}, \bibinfo{year}{2016}a.
\newblock \bibinfo{title}{Identity mappings in deep residual networks}.
\newblock \href{http://arxiv.org/abs/1603.05027}{{\tt arXiv:1603.05027}}.
\bibitem[{He et~al.(2016b)He, Zhang, Ren and Sun}]{resnet152v2}
\bibinfo{author}{He, K.}, \bibinfo{author}{Zhang, X.}, \bibinfo{author}{Ren,
  S.}, \bibinfo{author}{Sun, J.}, \bibinfo{year}{2016}b.
\newblock \bibinfo{title}{Identity mappings in deep residual networks}.
\newblock \href{http://arxiv.org/abs/1603.05027}{{\tt arXiv:1603.05027}}.
\bibitem[{Hu et~al.(2022a)Hu, Wang, Wan, Bao and Zeng}]{hu2022uav}
\bibinfo{author}{Hu, G.}, \bibinfo{author}{Wang, T.}, \bibinfo{author}{Wan,
  M.}, \bibinfo{author}{Bao, W.}, \bibinfo{author}{Zeng, W.},
  \bibinfo{year}{2022}a.
\newblock \bibinfo{title}{Uav remote sensing monitoring of pine forest diseases
  based on improved mask r-cnn}.
\newblock \bibinfo{journal}{International Journal of Remote Sensing}
  \bibinfo{volume}{43}, \bibinfo{pages}{1274--1305}.
\bibitem[{Hu et~al.(2022b)Hu, Yao, Wan, Bao and Zeng}]{hu2022detection}
\bibinfo{author}{Hu, G.}, \bibinfo{author}{Yao, P.}, \bibinfo{author}{Wan, M.},
  \bibinfo{author}{Bao, W.}, \bibinfo{author}{Zeng, W.}, \bibinfo{year}{2022}b.
\newblock \bibinfo{title}{Detection and classification of diseased pine trees
  with different levels of severity from uav remote sensing images}.
\newblock \bibinfo{journal}{Ecological Informatics} \bibinfo{volume}{72},
  \bibinfo{pages}{101844}.
\bibitem[{Huang et~al.(2017a)Huang, Liu, van~der Maaten and
  Weinberger}]{densenet121}
\bibinfo{author}{Huang, G.}, \bibinfo{author}{Liu, Z.},
  \bibinfo{author}{van~der Maaten, L.}, \bibinfo{author}{Weinberger, K.Q.},
  \bibinfo{year}{2017}a.
\newblock \bibinfo{title}{Densely connected convolutional networks}.
\newblock \href{http://arxiv.org/abs/1608.06993}{{\tt arXiv:1608.06993}}.
\bibitem[{Huang et~al.(2017b)Huang, Liu, van~der Maaten and
  Weinberger}]{densenet201}
\bibinfo{author}{Huang, G.}, \bibinfo{author}{Liu, Z.},
  \bibinfo{author}{van~der Maaten, L.}, \bibinfo{author}{Weinberger, K.Q.},
  \bibinfo{year}{2017}b.
\newblock \bibinfo{title}{Densely connected convolutional networks}.
\newblock \href{http://arxiv.org/abs/1608.06993}{{\tt arXiv:1608.06993}}.
\bibitem[{Ibrahim et~al.(2008)Ibrahim, Razak, Yakub, Yusoff, Idris and
  Tamil}]{ibrahim2008quranic}
\bibinfo{author}{Ibrahim, N.J.}, \bibinfo{author}{Razak, Z.},
  \bibinfo{author}{Yakub, M.}, \bibinfo{author}{Yusoff, Z.M.},
  \bibinfo{author}{Idris, M.Y.I.}, \bibinfo{author}{Tamil, E.},
  \bibinfo{year}{2008}.
\newblock \bibinfo{title}{Quranic verse recitation feature extraction using
  mel-frequency cepstral coefficients (mfcc)}, in: \bibinfo{booktitle}{Proc.
  the 4th IEEE Int. Colloquium on Signal Processing and its Application
  (CSPA)}, \bibinfo{organization}{Citeseer}.
\bibitem[{{\.I}nik(2023)}]{inik2023cnn}
\bibinfo{author}{{\.I}nik, {\"O}.}, \bibinfo{year}{2023}.
\newblock \bibinfo{title}{Cnn hyper-parameter optimization for environmental
  sound classification}.
\newblock \bibinfo{journal}{Applied Acoustics} \bibinfo{volume}{202},
  \bibinfo{pages}{109168}.
\bibitem[{Khan et~al.(2022)Khan, Quadri, Banday and Shah}]{khan2022deep}
\bibinfo{author}{Khan, A.I.}, \bibinfo{author}{Quadri, S.},
  \bibinfo{author}{Banday, S.}, \bibinfo{author}{Shah, J.L.},
  \bibinfo{year}{2022}.
\newblock \bibinfo{title}{Deep diagnosis: A real-time apple leaf disease
  detection system based on deep learning}.
\newblock \bibinfo{journal}{Computers and Electronics in Agriculture}
  \bibinfo{volume}{198}, \bibinfo{pages}{107093}.
\bibitem[{Koubaa et~al.(2020)Koubaa, Aldawood, Saeed, Hadid, Ahmed, Saad,
  Alkhouja, Ammar and Alkanhal}]{koubaa2020smart}
\bibinfo{author}{Koubaa, A.}, \bibinfo{author}{Aldawood, A.},
  \bibinfo{author}{Saeed, B.}, \bibinfo{author}{Hadid, A.},
  \bibinfo{author}{Ahmed, M.}, \bibinfo{author}{Saad, A.},
  \bibinfo{author}{Alkhouja, H.}, \bibinfo{author}{Ammar, A.},
  \bibinfo{author}{Alkanhal, M.}, \bibinfo{year}{2020}.
\newblock \bibinfo{title}{Smart palm: An iot framework for red palm weevil
  early detection}.
\newblock \bibinfo{journal}{Agronomy} \bibinfo{volume}{10},
  \bibinfo{pages}{987}.
\bibitem[{Lütkebohle(2018)}]{rpwfao}
\bibinfo{author}{Lütkebohle, I.}, \bibinfo{year}{2018}.
\newblock \bibinfo{title}{{Food and Agriculture Organization (FAO)}}.
\newblock
  \bibinfo{howpublished}{\url{https://www.fao.org/neareast/news/view/en/c/1104475/}}.
\newblock \bibinfo{note}{[Online; accessed 31-October-2022]}.
\bibitem[{Mallick et~al.(2023)Mallick, Biswas, Das, Saha, Chakrabarti and
  Deb}]{mallick2023deep}
\bibinfo{author}{Mallick, M.T.}, \bibinfo{author}{Biswas, S.},
  \bibinfo{author}{Das, A.K.}, \bibinfo{author}{Saha, H.N.},
  \bibinfo{author}{Chakrabarti, A.}, \bibinfo{author}{Deb, N.},
  \bibinfo{year}{2023}.
\newblock \bibinfo{title}{Deep learning based automated disease detection and
  pest classification in indian mung bean}.
\newblock \bibinfo{journal}{Multimedia Tools and Applications}
  \bibinfo{volume}{82}, \bibinfo{pages}{12017--12041}.
\bibitem[{Mohamed et~al.(2021)Mohamed, Hany, Adly, Atwa and Ragai}]{9686081}
\bibinfo{author}{Mohamed, A.}, \bibinfo{author}{Hany, A.},
  \bibinfo{author}{Adly, I.}, \bibinfo{author}{Atwa, A.},
  \bibinfo{author}{Ragai, H.}, \bibinfo{year}{2021}.
\newblock \bibinfo{title}{Ai for acoustic early detection of the red palm
  weevil}, in: \bibinfo{booktitle}{2021 16th International Conference on
  Computer Engineering and Systems (ICCES)}, pp. \bibinfo{pages}{1--4}.
\bibitem[{Parvathy et~al.(2021)Parvathy, Jayan, Pathrose and
  Rajesh}]{parvathy2021convolutional}
\bibinfo{author}{Parvathy, S.}, \bibinfo{author}{Jayan, D.P.},
  \bibinfo{author}{Pathrose, N.}, \bibinfo{author}{Rajesh, K.},
  \bibinfo{year}{2021}.
\newblock \bibinfo{title}{Convolutional autoencoder based deep learning model
  for identification of red palm weevil signals}, in: \bibinfo{booktitle}{2021
  Asia-Pacific Signal and Information Processing Association Annual Summit and
  Conference (APSIPA ASC)}, \bibinfo{organization}{IEEE}. pp.
  \bibinfo{pages}{1987--1992}.
\bibitem[{Pinhas et~al.(2008)Pinhas, Soroker, Hetzroni, Mizrach, Teicher and
  Goldberger}]{PINHAS2008131}
\bibinfo{author}{Pinhas, J.}, \bibinfo{author}{Soroker, V.},
  \bibinfo{author}{Hetzroni, A.}, \bibinfo{author}{Mizrach, A.},
  \bibinfo{author}{Teicher, M.}, \bibinfo{author}{Goldberger, J.},
  \bibinfo{year}{2008}.
\newblock \bibinfo{title}{Automatic acoustic detection of the red palm weevil}.
\newblock \bibinfo{journal}{Computers and Electronics in Agriculture}
  \bibinfo{volume}{63}, \bibinfo{pages}{131--139}.
\bibitem[{Putra et~al.(2022)Putra, Wijayanto and Chulafak}]{putra2022oil}
\bibinfo{author}{Putra, Y.C.}, \bibinfo{author}{Wijayanto, A.W.},
  \bibinfo{author}{Chulafak, G.A.}, \bibinfo{year}{2022}.
\newblock \bibinfo{title}{Oil palm trees detection and counting on microsoft
  bing maps very high resolution (vhr) satellite imagery and unmanned aerial
  vehicles (uav) data using image processing thresholding approach}.
\newblock \bibinfo{journal}{Ecological Informatics} \bibinfo{volume}{72},
  \bibinfo{pages}{101878}.
\bibitem[{Rehman et~al.(2022)Rehman, Shafique, Ghadi, Boulila, Jan, Gadekallu,
  Driss and Ahmad}]{rehman2022novel}
\bibinfo{author}{Rehman, M.U.}, \bibinfo{author}{Shafique, A.},
  \bibinfo{author}{Ghadi, Y.Y.}, \bibinfo{author}{Boulila, W.},
  \bibinfo{author}{Jan, S.U.}, \bibinfo{author}{Gadekallu, T.R.},
  \bibinfo{author}{Driss, M.}, \bibinfo{author}{Ahmad, J.},
  \bibinfo{year}{2022}.
\newblock \bibinfo{title}{A novel chaos-based privacy-preserving deep learning
  model for cancer diagnosis}.
\newblock \bibinfo{journal}{IEEE Transactions on Network Science and
  Engineering} \bibinfo{volume}{9}, \bibinfo{pages}{4322--4337}.
\bibitem[{Rigakis et~al.(2021)Rigakis, Potamitis, Tatlas, Potirakis and
  Ntalampiras}]{smartcities4010017}
\bibinfo{author}{Rigakis, I.}, \bibinfo{author}{Potamitis, I.},
  \bibinfo{author}{Tatlas, N.A.}, \bibinfo{author}{Potirakis, S.M.},
  \bibinfo{author}{Ntalampiras, S.}, \bibinfo{year}{2021}.
\newblock \bibinfo{title}{Treevibes: Modern tools for global monitoring of
  trees for borers}.
\newblock \bibinfo{journal}{Smart Cities} \bibinfo{volume}{4},
  \bibinfo{pages}{271--285}.
\newblock \URLprefix \url{https://www.mdpi.com/2624-6511/4/1/17},
  \DOIprefix\doi{10.3390/smartcities4010017}.
\bibitem[{Roopaei et~al.(2017)Roopaei, Rad and Choo}]{roopaei2017cloud}
\bibinfo{author}{Roopaei, M.}, \bibinfo{author}{Rad, P.},
  \bibinfo{author}{Choo, K.K.R.}, \bibinfo{year}{2017}.
\newblock \bibinfo{title}{Cloud of things in smart agriculture: Intelligent
  irrigation monitoring by thermal imaging}.
\newblock \bibinfo{journal}{IEEE Cloud computing} \bibinfo{volume}{4},
  \bibinfo{pages}{10--15}.
\bibitem[{Sandler et~al.(2018)Sandler, Howard, Zhu, Zhmoginov and
  Chen}]{mobilenetv2}
\bibinfo{author}{Sandler, M.}, \bibinfo{author}{Howard, A.},
  \bibinfo{author}{Zhu, M.}, \bibinfo{author}{Zhmoginov, A.},
  \bibinfo{author}{Chen, L.C.}, \bibinfo{year}{2018}.
\newblock \bibinfo{title}{Mobilenetv2: Inverted residuals and linear
  bottlenecks}.
\newblock \href{http://arxiv.org/abs/1801.04381}{{\tt arXiv:1801.04381}}.
\bibitem[{Simonyan and Zisserman(2014a)}]{vgg16}
\bibinfo{author}{Simonyan, K.}, \bibinfo{author}{Zisserman, A.},
  \bibinfo{year}{2014}a.
\newblock \bibinfo{title}{Very deep convolutional networks for large-scale
  image recognition}.
\newblock \href{http://arxiv.org/abs/1409.1556}{{\tt arXiv:1409.1556}}.
\bibitem[{Simonyan and Zisserman(2014b)}]{vgg19}
\bibinfo{author}{Simonyan, K.}, \bibinfo{author}{Zisserman, A.},
  \bibinfo{year}{2014}b.
\newblock \bibinfo{title}{Very deep convolutional networks for large-scale
  image recognition}.
\newblock \href{http://arxiv.org/abs/1409.1556}{{\tt arXiv:1409.1556}}.
\bibitem[{Sinha and Dhanalakshmi(2022)}]{sinha2022recent}
\bibinfo{author}{Sinha, B.B.}, \bibinfo{author}{Dhanalakshmi, R.},
  \bibinfo{year}{2022}.
\newblock \bibinfo{title}{Recent advancements and challenges of internet of
  things in smart agriculture: A survey}.
\newblock \bibinfo{journal}{Future Generation Computer Systems}
  \bibinfo{volume}{126}, \bibinfo{pages}{169--184}.
\bibitem[{Su et~al.(2023)Su, Li, Zhang, Wu and Wang}]{su2023robust}
\bibinfo{author}{Su, Z.}, \bibinfo{author}{Li, M.}, \bibinfo{author}{Zhang,
  G.}, \bibinfo{author}{Wu, Q.}, \bibinfo{author}{Wang, Y.},
  \bibinfo{year}{2023}.
\newblock \bibinfo{title}{Robust audio copy-move forgery detection on short
  forged slices using sliding window}.
\newblock \bibinfo{journal}{Journal of Information Security and Applications}
  \bibinfo{volume}{75}, \bibinfo{pages}{103507}.
\bibitem[{Szegedy et~al.(2015)Szegedy, Vanhoucke, Ioffe, Shlens and
  Wojna}]{inceptionv3}
\bibinfo{author}{Szegedy, C.}, \bibinfo{author}{Vanhoucke, V.},
  \bibinfo{author}{Ioffe, S.}, \bibinfo{author}{Shlens, J.},
  \bibinfo{author}{Wojna, Z.}, \bibinfo{year}{2015}.
\newblock \bibinfo{title}{Rethinking the inception architecture for computer
  vision}.
\bibitem[{Toffa and Mignotte(2020)}]{toffa2020environmental}
\bibinfo{author}{Toffa, O.K.}, \bibinfo{author}{Mignotte, M.},
  \bibinfo{year}{2020}.
\newblock \bibinfo{title}{Environmental sound classification using local binary
  pattern and audio features collaboration}.
\newblock \bibinfo{journal}{IEEE Transactions on Multimedia}
  \bibinfo{volume}{23}, \bibinfo{pages}{3978--3985}.
\bibitem[{Wang et~al.(2021)Wang, Mao, Ashry, Al-Fehaid, Al-Shawaf, Ng, Yu and
  Ooi}]{s21051592}
\bibinfo{author}{Wang, B.}, \bibinfo{author}{Mao, Y.}, \bibinfo{author}{Ashry,
  I.}, \bibinfo{author}{Al-Fehaid, Y.}, \bibinfo{author}{Al-Shawaf, A.},
  \bibinfo{author}{Ng, T.K.}, \bibinfo{author}{Yu, C.}, \bibinfo{author}{Ooi,
  B.S.}, \bibinfo{year}{2021}.
\newblock \bibinfo{title}{Towards detecting red palm weevil using machine
  learning and fiber optic distributed acoustic sensing}.
\newblock \bibinfo{journal}{Sensors} \bibinfo{volume}{21}.
\newblock \URLprefix \url{https://www.mdpi.com/1424-8220/21/5/1592},
  \DOIprefix\doi{10.3390/s21051592}.
\bibitem[{Yang et~al.(2021)Yang, Shu, Chen, Ferrag, Wu, Nurellari and
  Huang}]{yang2021survey}
\bibinfo{author}{Yang, X.}, \bibinfo{author}{Shu, L.}, \bibinfo{author}{Chen,
  J.}, \bibinfo{author}{Ferrag, M.A.}, \bibinfo{author}{Wu, J.},
  \bibinfo{author}{Nurellari, E.}, \bibinfo{author}{Huang, K.},
  \bibinfo{year}{2021}.
\newblock \bibinfo{title}{A survey on smart agriculture: Development modes,
  technologies, and security and privacy challenges}.
\newblock \bibinfo{journal}{IEEE/CAA Journal of Automatica Sinica}
  \bibinfo{volume}{8}, \bibinfo{pages}{273--302}.
\bibitem[{Yu et~al.(2023)Yu, Xie and Huang}]{yu2023inception}
\bibinfo{author}{Yu, S.}, \bibinfo{author}{Xie, L.}, \bibinfo{author}{Huang,
  Q.}, \bibinfo{year}{2023}.
\newblock \bibinfo{title}{Inception convolutional vision transformers for plant
  disease identification}.
\newblock \bibinfo{journal}{Internet of Things} \bibinfo{volume}{21},
  \bibinfo{pages}{100650}.
\bibitem[{Zheng et~al.(2001)Zheng, Zhang and Song}]{zheng2001comparison}
\bibinfo{author}{Zheng, F.}, \bibinfo{author}{Zhang, G.},
  \bibinfo{author}{Song, Z.}, \bibinfo{year}{2001}.
\newblock \bibinfo{title}{Comparison of different implementations of mfcc}.
\newblock \bibinfo{journal}{Journal of Computer science and Technology}
  \bibinfo{volume}{16}, \bibinfo{pages}{582--589}.
\bibitem[{Zhu et~al.(2023)Zhu, Xie, Chen, Tan, Deng, Zheng, Hu, Mustafa, Chen,
  Yi et~al.}]{zhu2023knowledge}
\bibinfo{author}{Zhu, D.}, \bibinfo{author}{Xie, L.}, \bibinfo{author}{Chen,
  B.}, \bibinfo{author}{Tan, J.}, \bibinfo{author}{Deng, R.},
  \bibinfo{author}{Zheng, Y.}, \bibinfo{author}{Hu, Q.},
  \bibinfo{author}{Mustafa, R.}, \bibinfo{author}{Chen, W.},
  \bibinfo{author}{Yi, S.}, et~al., \bibinfo{year}{2023}.
\newblock \bibinfo{title}{Knowledge graph and deep learning based pest
  detection and identification system for fruit quality}.
\newblock \bibinfo{journal}{Internet of Things} \bibinfo{volume}{21},
  \bibinfo{pages}{100649}.

\end{thebibliography}



\end{document}